\pdfoutput=1

\documentclass[11pt]{article}

\usepackage[final]{acl}

\usepackage{xspace}
\usepackage{graphicx}
\usepackage{enumitem}
\usepackage{xcolor}
\usepackage{tikz-dependency}
\usetikzlibrary{shapes,arrows}
\usepackage{amssymb}
\usepackage{float}
\usepackage{todonotes}

\usepackage{hyperref}       
\usepackage{url}            
\usepackage{booktabs}       
\usepackage{amsfonts}       
\usepackage{nicefrac}       
\usepackage{doi}

\usepackage{caption}
\usepackage{subcaption}

\usepackage{lscape}
\usepackage{longtable} 
\usepackage{multirow}
\usepackage{dsfont}

\setlist[description]{leftmargin=\parindent,labelindent=0pt}

\newcommand{\orange}[1]{\textcolor{black!20!orange}{#1}}
\newcommand{\purple}[1]{\textcolor{purple}{#1}}
\newcommand{\darkgreen}[1]{\textcolor{black!20!green}{#1}}

\newcommand{\enquote}[1]{``#1''}

\definecolor{talita}{rgb}{0.635,0.998,0.722}
\definecolor{anne}{rgb}{0.8,0.8,1}
\definecolor{sophie}{rgb}{0.998,0.722,0.635}
\definecolor{heike}{rgb}{0.4, 0.8, 0.4}
\definecolor{jakob}{rgb}{0.8,0.4,0.7}
\definecolor{steffen}{rgb}{0.2,0.4,0.8}
\definecolor{final}{rgb}{1, 1, 0.6}

\usepackage{color}
\usepackage{soul}

\newcounter{example}

\usepackage{array}
\newcolumntype{L}[1]{>{\raggedright\let\newline\\\arraybackslash\hspace{0pt}}m{#1}}
\newcolumntype{C}[1]{>{\centering\let\newline\\\arraybackslash\hspace{0pt}}m{#1}}
\newcolumntype{R}[1]{>{\raggedleft\let\newline\\\arraybackslash\hspace{0pt}}m{#1}}
\usepackage{rotating}

\usepackage{times}
\usepackage{latexsym}

\usepackage[T1]{fontenc}

\usepackage[utf8]{inputenc}

\usepackage{microtype}

\usepackage{inconsolata}

\usepackage{graphicx}

\usepackage[capitalise]{cleveref}
\usepackage{booktabs}
\usepackage{twemojis}

\crefformat{section}{section #2#1#3}
\crefformat{subsection}{section #2#1#3}
\crefformat{subsubsection}{section #2#1#3}
\crefmultiformat{section}{sections #2#1#3}{ and~#2#1#3}{, #2#1#3}{, and~#2#1#3}
\crefmultiformat{subsection}{sections #2#1#3}{ and~#2#1#3}{, #2#1#3}{, and~#2#1#3}
\crefmultiformat{subsubsection}{sections #2#1#3}{ and~#2#1#3}{, #2#1#3}{, and~#2#1#3}
\crefrangeformat{section}{\mbox{sections #3#1#4--#5#2#6}}
\crefrangeformat{subsection}{\mbox{sections #3#1#4--#5#2#6}}
\crefrangeformat{subsubsection}{\mbox{sections #3#1#4--#5#2#6}}

\title{OMoS-QA: A Dataset for Cross-Lingual Extractive\\ Question Answering in a German Migration Context}

\author{Steffen Kleinle\textsuperscript{1,2} \hspace*{7mm} Jakob Prange\textsuperscript{1} \hspace*{7mm} Annemarie Friedrich\textsuperscript{1} \\
 \textsuperscript{1}University of Augsburg, \textsuperscript{2}Tür an Tür Digitalfabrik GmbH\\
Contact: \{firstname.lastname\}@uni-a.de}

\begin{document}

\maketitle

\begin{abstract}
When immigrating to a new country, it is easy to feel overwhelmed by the need to obtain information on financial support, housing, schooling, language courses, and other issues.
If relocation is rushed or even forced, the necessity for high-quality answers to such questions is all the more urgent.
Official immigration counselors are usually overbooked, and online systems could guide newcomers to the requested information or a suitable counseling service.

To this end, we present OMoS-QA, a dataset of German and English questions paired with relevant trustworthy documents and manually annotated answers, specifically tailored to this scenario.
Questions are
automatically generated with an open-weights large language model (LLM) and answer sentences are
selected by crowd workers with high agreement.
With our data, we conduct a comparison of 5 pretrained LLMs on the task of extractive question answering (QA) in German and English. Across all models and both languages, we find high precision and low-to-mid recall in selecting answer sentences, which is a favorable trade-off to avoid misleading users. This performance even holds up when the question language does not match the document language. When it comes to identifying unanswerable questions given a context, there are larger differences between the two languages.
\end{abstract}

\section{Introduction}

Access to information is vital when moving to a new country, especially if the relocation is forced upon a person by war or persecution. 
Not knowing how to navigate immigration procedures and daily life in the host country can lead not only to confusion, insecurities, and delayed integration, but even to homelessness or deportation.
NLP methods can and should be used to critically analyze public-policy \citep{beese-etal-2022-fairger,blaette-etal-2020-europeanization1}  and general-public discourse about immigration \citep{wang-2024-metaphorical,lapesa-etal-2020-debatenet,sanguinetti-etal-2018-italian,ross-etal-2016-measuring}, to help newcomers learn new languages
\citep[][\textit{inter alia}]{bea-2023-innovative,nlp4call-2023-natural},
and to provide answers to their every\-day and immigration-related questions across languages and topics (this work).

\begin{figure}
    \centering
    \includegraphics[width=\columnwidth]{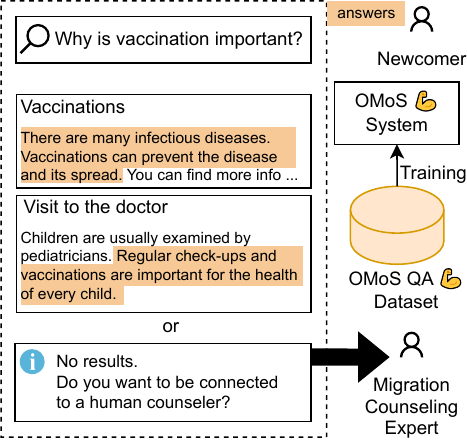}
    \caption{Overview of our proposed task, system, and new dataset, OMoS-QA \twemoji{1f4aa}: After the user asks a question, the system retrieves relevant documents and extracts answer sentences. The system is evaluated using the OMoS-QA \twemoji{1f4aa} corpus.}
    \label{fig:teaser}
\end{figure}

\begin{figure*}[ht]
    \centering
    \includegraphics[width=0.951\textwidth]{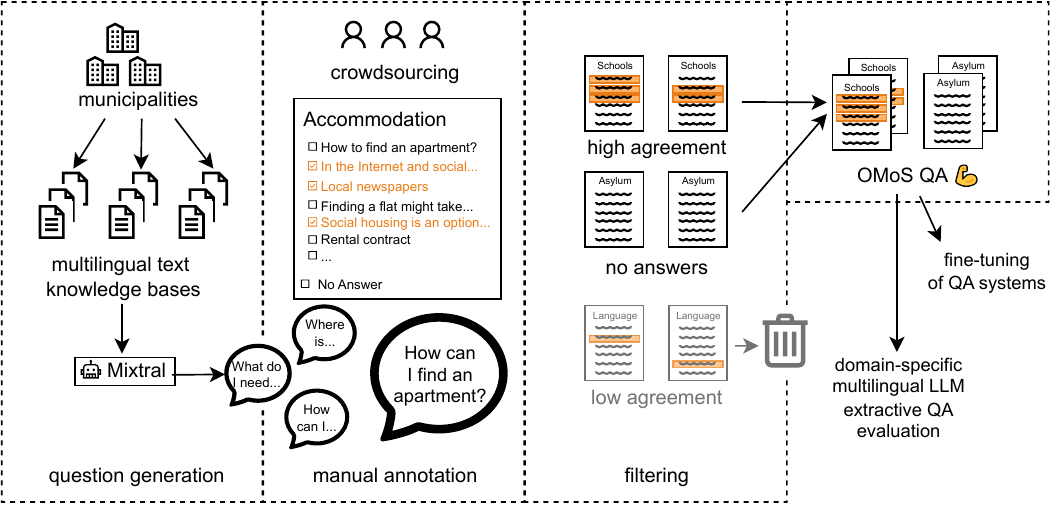}
    \caption{\textbf{OMoS-QA dataset creation.} Documents are taken from real-life multilingual knowledge bases. Questions are generated using Mixtral, but answers are annotated manually using crowdsourcing. The double-annotated dataset is then filtered on a question-level according to inter-annotator agreement.}
    \label{fig:dataset-creation}
\end{figure*}

In this paper, we address the latter issue by presenting OMoS-QA,\footnote{\textit{German:} \textbf{O}nline \textbf{M}igrationsberatung \textbf{o}hne \textbf{S}prachbarrieren; \textit{English:} Online migration counseling without language barriers. Data and code available at \href{https://github.com/digitalfabrik/integreat-qa-dataset}{https://github.com/digitalfabrik/integreat-qa-dataset}.\\ ``omos'' is also Greek for ``shoulder with upper arm'' \texttwemoji{1f4aa}.} an extractive QA dataset designed to support the development and rigorous testing of an online counseling system.
We envision an application-tailored multilingual question-answering (QA) system which, given a question and a collection of informative and instructive texts, identifies sentences providing evidence for answering the question in a relevant document (\cref{fig:teaser}).

Germany has seen multiple waves of immigration since the 1950s, most recently more than one million war refugees from Syria, Iraq, and Afghanistan since 2015 and around one million war refugees from Ukraine since 2022 and ongoing.
The German social system, aiming to support them, is known to be progressive
but at the same time bureaucratic.\footnote{For example, there is a law that regulates who may or may not provide official immigration counseling.}
Providing the necessary customized information to each individual is an enormous logistical challenge.
In particular during sudden crises,
the counseling system has insufficient personnel capacities to sustain one-on-one counseling for less urgent inquiries.
Hence, online resources are provided by cities and state governments, as well as NGOs.
However, online information is scattered across many websites and portals, where it is location-specific, unstructured or structured inconsistently, and needs to be updated periodically---all on top of the language barrier.

OMoS-QA treats QA as a sentence extraction task rather than text generation, because faithfulness is of utmost importance. Well-known risks of free-text generation with large language models (LLMs), such as made-up facts and hallucinated entities \citep{shah-bender-2024-envisioning,ji-etal-2023-hallucination,mckenna-etal-2023-sources}, are not acceptable in our application scenario of supporting migrants with information about social, economic, and legal processes.
For the same reason, our approach aims to detect if a question is unanswerable given the provided evidence context.
Extracting full sentences rather than token spans further helps with completeness and readability of the answers shown to the user.
The process for constructing our new dataset is illustrated in \cref{fig:dataset-creation}.
The contributions of this work are as follows.

\begin{itemize}
\setlength\itemsep{0pt}
    \item We present OMoS-QA, a manually annotated \textbf{corpus} of questions in German and English paired with relevant informational documents about a variety of social, economic, and legal topics and support offers. The documents were provided by three German municipalities, questions were \textbf{generated with an open-weight large language model} (LLM),
    and answer annotations were collected via voluntary \textbf{crowd-sourcing} (\cref{sec:corpus}).
    \item In order to construct a high quality dataset from the crowd-sourced annotations, we develop a filtering method based on a chance-corrected version of the Jaccard coefficient. We also present a detailed inter-annotator agreement study.
    \item Finally, we experiment with state-of-the-art pretrained LLMs (\cref{sec:experiments}).
    We compare 4 open-weight models as well as GPT-3.5, finding overall high precision in answer sentence selection and high recall in identifying unanswerable questions.
    A pilot cross-language QA study yields promising results.
\end{itemize}

\clearpage

\section{Related Work}\label{sec:relwork}

To ensure faithfulness of responses in our highly sensitive socio-political scenario, we focus exclusively on \textbf{extractive QA}, where the model is given a specific context to read, from which it should extract answers.
\citet{luo-etal-2022-choose} provide a helpful comparative overview of extractive and generative approaches, and \citet{luthier-popescu-belis-2020-chat} have shown advantages of a hybrid system which dynamically chooses one of the two strategies.

Below we discuss related work on QA dataset construction, modeling extractive QA, and further NLP research in similar socio-political contexts.

\paragraph{QA Dataset Construction.}
The most popular QA datasets, such as SQuAD \citep{rajpurkar-etal-2016-squad} and its derivatives \citep[e.g.][]{rajpurkar-etal-2018-know,moller-etal-2021-germanquad}, are general-purpose and thus not directly applicable to our scenario.
However, curating and annotating a new QA corpus requires some finesse, especially when the target application is highly task-specific \citep{agarwal-etal-2022-knowledge,xu-etal-2022-task} or lies in a specific domain \citep{bechet-etal-2022-question,han-etal-2022-generating}.

There is some consensus that \textbf{question generation} (QG) can be mostly automated, whereas ground-truth \textbf{answer annotations} should be provided by humans to ensure correctness.
QG techniques that have proven useful include using a short summary of the context as input to the QG model \citep{dugan-etal-2022-feasibility}; question rewriting \citep{brabant-etal-2022-coqar}; running  QA as an auxiliary task and rewarding consistency between questions and answers \citep{yuan-etal-2023-selecting,dugan-etal-2022-feasibility}; extracting QA-pairs from video transcripts \citep{westera-etal-2020-ted,pouran-ben-veyseh-etal-2022-behanceqa}; and prompt engineering towards quality and diversity of the generated sentences \citep{schick-schutze-2021-generating,yuan-etal-2023-selecting}.
Manual answer annotation via crowd-sourcing, particularly making QA and other NLP tasks such as semantic role labeling (SRL) accessible to laypeople, has been popularized by the QA-SRL project \citep{he-etal-2015-question,roit-etal-2020-controlled,brook-weiss-etal-2021-qa}.

In order to maintain high precision, we are particularly concerned with the option of marking a question as \textbf{unanswerable} given a context \citep[cf.][]{rajpurkar-etal-2018-know,liu-etal-2020-roberta,henning-etal-2023-answer}. Moreover, \citet{lauriola-etal-2022-building} have built a dataset of questions requiring clarifications, which we will consider in future work.

Finally, while \textbf{multi- and cross-linguality} remains a major challenge \citep{charlet-etal-2020-cross}, QA datasets in many languages (besides English) have been created in recent years, for German most notably by \citet{moller-etal-2021-germanquad}.

\paragraph{Extractive QA Modeling.}

Approaches to extractive QA vary in whether they aim to predict a single span of a few tokens \citep{seo-etal-2017-bidirectional,clark-gardner-2018-simple,hu-etal-2018-reinforced}, or whether the aim is to collect supporting evidence for a (possibly latent) answer \citep{murdock-etal-2012-textual}.
To extract evidence sentences for choosing an answer in a multiple-choice QA setting, \citet{wang-etal-2019-evidence} finetune a GPT model \citep{radford-etal-2018-improving}.
\citet{narayan-etal-2018-document} model the whole document via LSTMs over sentences before choosing sentences for answer selection and extractive summarization.
\citet{yoon-etal-2020-propagate} detect sentences for answering multi-hop questions with a graph neural net-based model that also takes the passage structure of the context into account.

Perhaps the closest to our problem setting in that both unanswered questions and discontiguous multi-span responses need to be accounted for (albeit in different application scenarios) are the works of \citet{prasad-etal-2023-meetingqa} and \citet{henning-etal-2023-answer}.
\citeauthor{prasad-etal-2023-meetingqa} compare several pretrained BERT-style models in a multi-turn dialog setting while \citeauthor{henning-etal-2023-answer} prompt a generative model to extract sentence numbers to answer questions on instructive texts.

\paragraph{Socio-political NLP Applications.}

In order to track, analyze, and predict trends in parliamentary debates about migrants and migration, \citet{blaette-etal-2020-europeanization1} employ topic models while \citet{beese-etal-2022-fairger} finetune a BERT model.
A number of corpora have been compiled to study the public debate about immigration-related questions in Europe: e.g., in German and Slovene news \citep{lapesa-etal-2020-debatenet,zwitter-vitez-etal-2022-extracting}, German and Italian social media \citep{ross-etal-2016-measuring,sanguinetti-etal-2018-italian}, and UK partisan media \citep{wang-2024-metaphorical}.

\section{The OMoS-QA Corpus}\label{sec:corpus}

In this work, we present OMoS-QA, a novel dataset for QA in the context of \underline{O}nline \underline{M}igrationsberatung \underline{o}hne \underline{S}prachbarrieren (online migration counseling without language barriers). In its current version, it consists of over 900 automatically generated questions and manual answer annotations on documents contextually relevant to our problem setting in both German and English.
In this section, we describe the dataset collection, annotation, and filtering, and provide corpus statistics.

\begin{table*}[ht]
    \centering
    \footnotesize
    \begin{tabular}{llrrrr}
\toprule
  &  & \multicolumn{1}{c}{train} & \multicolumn{1}{c}{dev} & \multicolumn{1}{c}{test} & \multicolumn{1}{c}{total} \\
\midrule
 German & Questions & 338\phantom{.00 {\scriptsize $\pm$ 00.00}} & 143\phantom{.00 {\scriptsize $\pm$ 00.00}} & 185\phantom{.00 {\scriptsize $\pm$ 00.00}} & 666\phantom{.00 {\scriptsize $\pm$ 00.00}} \\
  & \hspace{1em}No Answer & 63 \textcolor{gray}{\scriptsize (19\%)}\phantom{.00 \scriptsize ..} & 30 \textcolor{gray}{\scriptsize (21\%)}\phantom{.00 \scriptsize ..} & 43 \textcolor{gray}{\scriptsize (23\%)}\phantom{.00 \scriptsize ..} & 136 \textcolor{gray}{\scriptsize (20\%)}\phantom{.00 \scriptsize ..} \\
  & \hspace{1em}Contiguous Answer & 209 \textcolor{gray}{\scriptsize (62\%)}\phantom{.00 \scriptsize ..} & 86 \textcolor{gray}{\scriptsize (60\%)}\phantom{.00 \scriptsize ..} & 104 \textcolor{gray}{\scriptsize (56\%)}\phantom{.00 \scriptsize ..} & 399 \textcolor{gray}{\scriptsize (60\%)}\phantom{.00 \scriptsize ..} \\
  & \hspace{1em}Non-Contiguous Answer & 66 \textcolor{gray}{\scriptsize (20\%)}\phantom{.00 \scriptsize ..} & 27 \textcolor{gray}{\scriptsize (19\%)}\phantom{.00 \scriptsize ..} & 38 \textcolor{gray}{\scriptsize (21\%)}\phantom{.00 \scriptsize ..} & 131 \textcolor{gray}{\scriptsize (20\%)}\phantom{.00 \scriptsize ..} \\
  & Documents & 205\phantom{.00 {\scriptsize $\pm$ 00.00}} & 90\phantom{.00 {\scriptsize $\pm$ 00.00}} & 117\phantom{.00 {\scriptsize $\pm$ 00.00}} & 412\phantom{.00 {\scriptsize $\pm$ 00.00}} \\
  & Questions/Document & 1.65 \phantom{\scriptsize $\pm$ 00.00} & 1.59 \phantom{\scriptsize $\pm$ 00.00} & 1.58 \phantom{\scriptsize $\pm$ 00.00} & 1.62 \phantom{\scriptsize $\pm$ 00.00} \\
  & Sentences/Document & 27.16 \textcolor{gray}{\scriptsize $\pm$ 20.11} & 27.96 \textcolor{gray}{\scriptsize $\pm$ 15.87} & 26.91 \textcolor{gray}{\scriptsize $\pm$ 17.88} & 27.26 \textcolor{gray}{\scriptsize $\pm$ 18.59} \\
  & Chars/Sentence & 58.62 \textcolor{gray}{\scriptsize $\pm$ 15.93} & 61.74 \textcolor{gray}{\scriptsize $\pm$ 16.32} & 61.96 \textcolor{gray}{\scriptsize $\pm$ 17.25} & 60.25 \textcolor{gray}{\scriptsize $\pm$ 16.44} \\
  & Chars/Question & 57.85 \textcolor{gray}{\scriptsize $\pm$ 15.68} & 58.91 \textcolor{gray}{\scriptsize $\pm$ 17.21} & 59.61 \textcolor{gray}{\scriptsize $\pm$ 16.45} & 58.56 \textcolor{gray}{\scriptsize $\pm$ 16.23} \\
  & Agreement (Jaccard) & 0.60 \phantom{\scriptsize 0}\textcolor{gray}{\scriptsize $\pm$ 0.33} & 0.59 \phantom{\scriptsize 0}\textcolor{gray}{\scriptsize $\pm$ 0.33} & 0.60 \phantom{\scriptsize 0}\textcolor{gray}{\scriptsize $\pm$ 0.34} & 0.60 \phantom{\scriptsize 0}\textcolor{gray}{\scriptsize $\pm$ 0.33} \\
  & \hspace{1em}with adjacent sentences & 0.86 \phantom{\scriptsize 0}\textcolor{gray}{\scriptsize $\pm$ 0.19} & 0.85 \phantom{\scriptsize 0}\textcolor{gray}{\scriptsize $\pm$ 0.18} & 0.86 \phantom{\scriptsize 0}\textcolor{gray}{\scriptsize $\pm$ 0.19} & 0.86 \phantom{\scriptsize 0}\textcolor{gray}{\scriptsize $\pm$ 0.19} \\
  & Answer Sentences/Question & 5.37 \phantom{\scriptsize 0}\textcolor{gray}{\scriptsize $\pm$ 6.09} & 5.57 \phantom{\scriptsize 0}\textcolor{gray}{\scriptsize $\pm$ 5.89} & 5.29 \phantom{\scriptsize 0}\textcolor{gray}{\scriptsize $\pm$ 6.84} & 5.39 \phantom{\scriptsize 0}\textcolor{gray}{\scriptsize $\pm$ 6.26} \\
  & Answers Sentences/Total Sentences & 0.28 \phantom{\scriptsize 0}\textcolor{gray}{\scriptsize $\pm$ 0.29} & 0.25 \phantom{\scriptsize 0}\textcolor{gray}{\scriptsize $\pm$ 0.27} & 0.27 \phantom{\scriptsize 0}\textcolor{gray}{\scriptsize $\pm$ 0.28} & 0.27 \phantom{\scriptsize 0}\textcolor{gray}{\scriptsize $\pm$ 0.28} \\
\midrule
 English & Questions & 123\phantom{.00 {\scriptsize $\pm$ 00.00}} & 50\phantom{.00 {\scriptsize $\pm$ 00.00}} & 67\phantom{.00 {\scriptsize $\pm$ 00.00}} & 240\phantom{.00 {\scriptsize $\pm$ 00.00}} \\
  & \hspace{1em}No Answer & 18 \textcolor{gray}{\scriptsize (15\%)}\phantom{.00 \scriptsize ..} & 8 \textcolor{gray}{\scriptsize (16\%)}\phantom{.00 \scriptsize ..} & 12 \textcolor{gray}{\scriptsize (18\%)}\phantom{.00 \scriptsize ..} & 38 \textcolor{gray}{\scriptsize (16\%)}\phantom{.00 \scriptsize ..} \\
  & \hspace{1em}Contiguous Answer & 95 \textcolor{gray}{\scriptsize (77\%)}\phantom{.00 \scriptsize ..} & 38 \textcolor{gray}{\scriptsize (76\%)}\phantom{.00 \scriptsize ..} & 49 \textcolor{gray}{\scriptsize (73\%)}\phantom{.00 \scriptsize ..} & 182 \textcolor{gray}{\scriptsize (76\%)}\phantom{.00 \scriptsize ..} \\
  & \hspace{1em}Non-Contiguous Answer & 10 \phantom{\scriptsize 0}\textcolor{gray}{\scriptsize (8\%)}\phantom{.00 \scriptsize ..} & 4 \phantom{\scriptsize 0}\textcolor{gray}{\scriptsize (8\%)}\phantom{.00 \scriptsize ..} & 6 \phantom{\scriptsize 0}\textcolor{gray}{\scriptsize (9\%)}\phantom{.00 \scriptsize ..} & 20 \phantom{\scriptsize 0}\textcolor{gray}{\scriptsize (8\%)}\phantom{.00 \scriptsize ..} \\
  & Documents & 103\phantom{.00 {\scriptsize $\pm$ 00.00}} & 43\phantom{.00 {\scriptsize $\pm$ 00.00}} & 59\phantom{.00 {\scriptsize $\pm$ 00.00}} & 205\phantom{.00 {\scriptsize $\pm$ 00.00}} \\
  & Questions/Document & 1.19 \phantom{\scriptsize $\pm$ 00.00} & 1.16 \phantom{\scriptsize $\pm$ 00.00} & 1.14 \phantom{\scriptsize $\pm$ 00.00} & 1.17 \phantom{\scriptsize $\pm$ 00.00} \\
  & Sentences/Document & 23.51 \textcolor{gray}{\scriptsize $\pm$ 13.30} & 25.58 \textcolor{gray}{\scriptsize $\pm$ 16.68} & 25.49 \textcolor{gray}{\scriptsize $\pm$ 13.68} & 24.52 \textcolor{gray}{\scriptsize $\pm$ 14.14} \\
  & Chars/Sentence & 65.28 \textcolor{gray}{\scriptsize $\pm$ 18.22} & 61.74 \textcolor{gray}{\scriptsize $\pm$ 12.72} & 60.48 \textcolor{gray}{\scriptsize $\pm$ 15.30} & 63.16 \textcolor{gray}{\scriptsize $\pm$ 16.45} \\
  & Chars/Question & 59.46 \textcolor{gray}{\scriptsize $\pm$ 15.98} & 56.48 \textcolor{gray}{\scriptsize $\pm$ 13.22} & 56.51 \textcolor{gray}{\scriptsize $\pm$ 14.72} & 58.01 \textcolor{gray}{\scriptsize $\pm$ 15.11} \\
    & Agreement (Jaccard) & 0.58 \phantom{\scriptsize 0}\textcolor{gray}{\scriptsize $\pm$ 0.34} & 0.59 \phantom{\scriptsize 0}\textcolor{gray}{\scriptsize $\pm$ 0.32} & 0.56 \phantom{\scriptsize 0}\textcolor{gray}{\scriptsize $\pm$ 0.34} & 0.58 \phantom{\scriptsize 0}\textcolor{gray}{\scriptsize $\pm$ 0.34} \\
  & \hspace{1em}with adjacent sentences & 0.86 \phantom{\scriptsize 0}\textcolor{gray}{\scriptsize $\pm$ 0.20} & 0.84 \phantom{\scriptsize 0}\textcolor{gray}{\scriptsize $\pm$ 0.19} & 0.86 \phantom{\scriptsize 0}\textcolor{gray}{\scriptsize $\pm$ 0.20} & 0.86 \phantom{\scriptsize 0}\textcolor{gray}{\scriptsize $\pm$ 0.19} \\
  & Answer Sentences/Question & 4.41 \phantom{\scriptsize 0}\textcolor{gray}{\scriptsize $\pm$ 4.98} & 3.90 \phantom{\scriptsize 0}\textcolor{gray}{\scriptsize $\pm$ 3.62} & 4.19 \phantom{\scriptsize 0}\textcolor{gray}{\scriptsize $\pm$ 4.39} & 4.24 \phantom{\scriptsize 0}\textcolor{gray}{\scriptsize $\pm$ 4.55} \\
  & Answers Sentences/Total Sentences & 0.23 \phantom{\scriptsize 0}\textcolor{gray}{\scriptsize $\pm$ 0.23} & 0.20 \phantom{\scriptsize 0}\textcolor{gray}{\scriptsize $\pm$ 0.21} & 0.22 \phantom{\scriptsize 0}\textcolor{gray}{\scriptsize $\pm$ 0.24} & 0.22 \phantom{\scriptsize 0}\textcolor{gray}{\scriptsize $\pm$ 0.23} \\
\midrule
 All & Questions & 461\phantom{.00 {\scriptsize $\pm$ 00.00}} & 193\phantom{.00 {\scriptsize $\pm$ 00.00}} & 252\phantom{.00 {\scriptsize $\pm$ 00.00}} & 906\phantom{.00 {\scriptsize $\pm$ 00.00}} \\
  & \hspace{1em}No Answer & 81 \textcolor{gray}{\scriptsize (18\%)}\phantom{.00 \scriptsize ..} & 38 \textcolor{gray}{\scriptsize (20\%)}\phantom{.00 \scriptsize ..} & 55 \textcolor{gray}{\scriptsize (22\%)}\phantom{.00 \scriptsize ..} & 174 \textcolor{gray}{\scriptsize (19\%)}\phantom{.00 \scriptsize ..} \\
  & \hspace{1em}Contiguous Answer & 304 \textcolor{gray}{\scriptsize (66\%)}\phantom{.00 \scriptsize ..} & 124 \textcolor{gray}{\scriptsize (64\%)}\phantom{.00 \scriptsize ..} & 153 \textcolor{gray}{\scriptsize (61\%)}\phantom{.00 \scriptsize ..} & 581 \textcolor{gray}{\scriptsize (64\%)}\phantom{.00 \scriptsize ..} \\
  & \hspace{1em}Non-Contiguous Answer & 76 \textcolor{gray}{\scriptsize (16\%)}\phantom{.00 \scriptsize ..} & 31 \textcolor{gray}{\scriptsize (16\%)}\phantom{.00 \scriptsize ..} & 44 \textcolor{gray}{\scriptsize (17\%)}\phantom{.00 \scriptsize ..} & 151 \textcolor{gray}{\scriptsize (17\%)}\phantom{.00 \scriptsize ..} \\
\bottomrule
    \end{tabular}
    \caption{OMoS-QA: Overview of corpus statistics of final dataset. The Jaccard index is chance-corrected.}
    \label{tab:corpus-stats}
\end{table*}

\subsection{Data Collection and Annotation}

In an initial attempt, we tried to elicit common questions and their answers from administrative staff of migration agencies and NGO volunteers. This was unsuccessful due to their limited availability and the substantial time requirements necessary for the task.
Therefore, inspired by \citet{schick-schutze-2021-generating}, we leverage the capabilities of LLMs to automatically generate questions.
To ensure a high quality of the dataset, we collect at least two human answer annotations per question, facilitated by a new custom annotation tool. Only annotations that are largely agreed upon by two annotators are included in the final dataset. 

\paragraph{Question Generation.}\label{sec:question-generation}
We used Mixtral-8x7B-Instruct-v0.1 \citep[henceforth abbreviated as Mixtral-8x7B;][]{jiang2024mixtral} to generate questions for German and English documents provided under CC BY 4.0 by three municipalities in Southern Germany.\footnote{The city of Munich and the districts (Landkreise) Augsburg and Rems-Murr-Kreis.} The documents were retrieved using the {Integreat API}\footnote{\url{https://digitalfabrik.github.io/integreat-cms/api-docs.html\#pages}} on 2024-02-02.
To facilitate the diversity of the dataset and to include both answerable and unanswerable questions, we employed two different question generation strategies for every document.
In the first, the prompt contained the full document, in the second, we only provided an automatically generated three-word summary.
The second strategy aimed at eliciting questions that are unanswerable given the provided document.

All questions were manually filtered, and in some cases corrected by the first author,
e.g., ``What are the emergency numbers provided?''~was edited to~``What emergency numbers are available?.''
In total, we collected 1,844 German questions for 548 documents and 3,062 English questions for 652 documents. Around 60\% of the questions have been generated from a three-word summary such as ``domestic violence support,'' ``refugee counseling services,'' or ``recognition of degrees.''

\paragraph{Human Annotations.}\label{sec:human-annotations}
The task of finding the answers within documents resided with human annotators.
As we resort to voluntary crowdsourcing, we aim to make the annotation process easy and time-efficient by creating a custom web-based annotation tool (see \cref{apx:annotation-tool}) tailored to our use case. 
We frame the annotation task as the selection of one or multiple complete sentences that help to answer the question.
Annotators are shown a question together with the text, and the option to select sentences via checkboxes.
If no answer is found in the text, a separate checkbox has to be selected to consciously confirm this decision. 

The annotators were recruited on a voluntary basis from German NGOs in the migration context and in the personal environment of the authors.
Questions are randomly assigned to annotators on-the-fly, allowing each person to do as many (or few) annotations as they want. In total, we gathered 3,688 annotations for 1,944 questions by 238 annotators.

\subsection{Question Filtering}
To account for voluntary or involuntary mistakes, biases, and subjective answers by annotators, we require two annotations per question by different annotators. The annotations therefore amount to 1,744 questions with two annotations (de: 1,268, en: 476) for 863 different documents.
To filter questions with low \textbf{inter-annotator agreement} (IAA), we measure question-level agreement using the \emph{Jaccard index} over the two sets of sentences judged as relevant to answering the question by the two annotators.
In a nutshell, the Jaccard index is defined as \enquote{intersection over union.}

For measuring agreement, we use a chance-corrected Jaccard index.
Our metric captures how much the two annotators agree on the selected set of sentences beyond chance.
We assume, admittedly over-simplifying, that the prior probability of selecting a sentence is independent of the question, document, and annotator, and compute it as the total fraction of sentence selections over two times the corpus size (as each document receives two annotations).
For details, see \cref{app:agreement-metric}.
In our case, $P(sel)$ is 0.1856, and the expected agreement amounts to a Jaccard index of only 0.0344.

\begin{figure}[t]
    \centering
    \includegraphics[width=0.8\columnwidth]{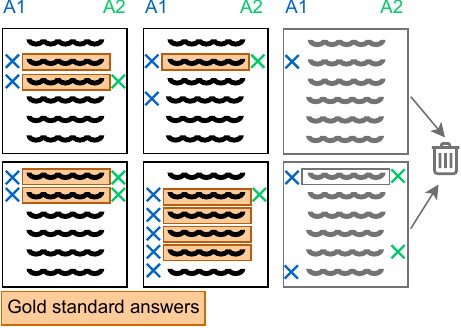}
    \caption{Gold standard construction from labels of two human annotators A1 (blue) and A2 (green). The gold standard contains sentences that A1 and A2 both mark as answers, as well as adjacent sentences marked by only one of them if at most three sentences away from the agreed-upon answer.}
    \label{fig:questionextension}
\end{figure}

The average IAA over all questions is 0.34 (chance corrected: 0.31). This can be partly attributed to the fact that most questions are non-factoid, i.e., answers are not objective single \enquote{facts} but instead one or more relevant sentences where the boundaries around what should be the core answer and what is additional context are difficult to draw.
To account for this difficulty, we modify the annotations in a heuristic way as illustrated in \cref{fig:questionextension}. For each sentence marked by just one of the annotators that is adjacent to a sentence marked as relevant by both annotators, we change the annotation of the respective other annotator to \enquote{relevant} as well.
We do this only if the sentence originally marked by both annotators is no more than three\footnote{This threshold is chosen as a middle ground between too little and too much additional context backed up by a manual inspection of samples.} sentences away.

After modifying the annotations to include adjacent sentences, the average Jaccard index is 0.50 (chance corrected: 0.48). To assure a high quality dataset, we filter out questions with a (non-chance-corrected) Jaccard index $<$0.5.
This leaves us with 906 (51\%) questions (de: 663, en: 243) with an average agreement of 0.86 (chance corrected: 0.86).
The agreement when leaving out the adjustment of including adjacent sentences amounts to 0.61 (chance corrected: 0.59).
As gold-standard answers we chose the intersection of both annotations, but including adjacent sentences as explained above.

\subsection{Final Dataset}\label{sec:final-data}
\cref{tab:corpus-stats} provides an overview of the corpus statistics of the final version of OMoS-QA.
Out of the 906 QA pairs included in our \textbf{final dataset}, 151 (16\%) have non-contiguous answers (i.e., the answer sentences are not adjacent), 110 (12\%) have a single answer sentence and 165 (18\%) questions have no answer in the document.
The IAA did not differ substantially between German and English annotations in both the raw dataset (de: 0.34, en: 0.32) as well as the final dataset (de, en: 0.86).

\textbf{Translations.}
To increase the size of the dataset and to take the multilingual setting into account, we translate the German questions and documents to English and vice versa using {DeepL}.\footnote{\url{https://developers.deepl.com/docs}}
In order to preserve the gold-standard answers represented by the sentence indices, we translate each context sentence-by-sentence. Accordingly, in the German version of the dataset 240 and in the English version 666 of the 906 questions are machine-translated.
We retain the information on the original languages.

\textbf{Dataset Split.}
We split our dataset into train (51\%), dev (21\%) and test (28\%) partitions with similar internal splits for the original language and the city the document is from. Questions without an answer, questions with contiguous and questions with non-contiguous answers are present with a similar probability over all partitions. As some questions refer to the same document, we make sure that no document occurs in multiple partitions.
The proposed split is assuring a close to uniform distribution of several key properties of the dataset such as the agreement of both annotations, the document length or the annotated answer count.

\begin{table*}[t]
    \setlength{\tabcolsep}{5.7pt}
    \centering
    \footnotesize
    \begin{tabular}{lc | ccc | ccc c ccc | ccc}
\toprule
\multicolumn{2}{c}{} & \multicolumn{6}{c}{Sentence-level Answers} & & \multicolumn{6}{c}{Question-level Unanswerability} \\
\cmidrule{1-8}\cmidrule{10-15} & & \multicolumn{3}{c|}{German} & \multicolumn{3}{c}{English} & & \multicolumn{3}{c|}{German} & \multicolumn{3}{c}{English} \\
 Model & Setting & P & R & F & P & R & F &  & P & R & F & P & R & F \\
\cmidrule{1 - 8}\cmidrule{10 - 15}
 Mixtral-8x7B & 0-shot & 74.5 & 47.1 & 57.7 & 73.4 & 44.2 & 55.2 & \textbf{} & 68.9 & 56.4 & 62.0 & 65.8 & 45.5 & 53.8 \\
  & 5-shot & 79.0 & 51.7 & \textbf{62.5} & 77.9 & 50.5 & 61.3 &  & 67.8 & 72.7 & 70.2 & 65.6 & 76.4 & 70.6 \\
\cmidrule{1 - 8}\cmidrule{10 - 15}
 Mistral-7B & 0-shot & 69.7 & 47.8 & 56.7 & 74.1 & 47.5 & 57.9 &  & \textbf{80.0} & 14.5 & 24.6 & 70.0 & 25.5 & 37.3 \\
  & 5-shot & \textbf{87.6} & 20.3 & 32.9 & 84.3 & 29.5 & 43.7 &  & 29.2 & \textbf{89.1} & 43.9 & 30.3 & 72.7 & 42.8 \\
\cmidrule{1 - 8}\cmidrule{10 - 15}
 Llama-3-8B & 0-shot & 74.9 & 30.0 & 42.9 & 78.2 & 34.8 & 48.1 &  & 71.1 & 49.1 & 58.1 & 54.7 & 52.7 & 53.7 \\
  & 5-shot & 81.9 & 42.2 & 55.7 & 82.1 & 44.2 & 57.4 &  & 54.7 & 85.5 & 66.7 & 53.6 & \textbf{81.8} & 64.7 \\
\cmidrule{1 - 8}\cmidrule{10 - 15}
 Llama-3-70B & 0-shot & 85.5 & 46.6 & 60.3 & 84.8 & 46.7 & 60.2 &  & 69.8 & 67.3 & 68.5 & 74.5 & 63.6 & 68.6 \\
  & 5-shot & 86.7 & 48.2 & 62.0 & 84.9 & 48.4 & 61.6 &  & 68.3 & 78.2 & \textbf{72.9} & 64.5 & 72.7 & 68.4 \\
\cmidrule{1 - 8}\cmidrule{10 - 15}
 GPT-3.5-Turbo & 0-shot & 85.3 & 31.6 & 46.1 & \textbf{87.3} & 31.2 & 45.9 &  & 50.8 & 60.0 & 55.0 & 54.4 & 67.3 & 60.2 \\
  & 5-shot & 81.8 & 45.1 & 58.1 & 83.8 & 43.9 & 57.6 &  & 70.9 & 70.9 & 70.9 & 67.2 & 74.5 & \textbf{70.7} \\
\cmidrule{1 - 8}\cmidrule{10 - 15}
 DeBERTa & $-$ & 62.6 & \textbf{62.4} & \textbf{62.5} & 65.7 & \textbf{64.2} & \textbf{64.9} &  & 56.2 & 65.5 & 60.5 & 59.4 & 69.1 & 63.9 \\
\cmidrule{1 - 8}\cmidrule{10 - 15}
\multicolumn{2}{l|}{\itshape Human Agreement*} & $-$ & $-$ & \textit{57.8} & $-$ & $-$ & \textit{57.8} &  & $-$ & $-$ & \textit{47.8} & $-$ & $-$ & \textit{47.8} \\
\multicolumn{2}{l|}{\textit{\hspace{6mm}test partition only}} & $-$ & $-$ & \textit{76.3} & $-$ & $-$ & \textit{76.3} &  & $-$ & $-$ & \textit{100.0} & $-$ & $-$ & \textit{100.0} \\ 
\bottomrule
    \end{tabular}
    \caption{Test set performance (in \%) of zero-shot and 5-shot LLMs and finetuned DeBERTa on sentence-level answer extraction (left) and detection of unanswerable questions (right). The best result in each column is \textbf{bolded}. \textit{*Human Agreement} is computed from agreement before the dataset filtering step (\cref{fig:dataset-creation}) and therefore not directly comparable to model performance.}
    \label{tab:qa-experiment-test}
\end{table*}

\section{Experiments}\label{sec:experiments}

In this section, we describe our experiments.
We evaluate several off-the-shelf LLMs as well as a finetuned sentence classifier on OMoS-QA.

\subsection{Setup}\label{sec:experiment-setup}
We mostly follow the \textbf{prompt templates} proposed by \citet{henning-etal-2023-answer} for both the 0-shot and 5-shot settings, instructing the models to output a list containing the sentence IDs of the answer sentences.\footnote{We used a model temperature of 0.75.}
We test the models in a 0-shot setting, only providing the prompt, but no concrete examples.
In addition, we test the models in a 5-shot setting, in which we manually select and chunk examples for both German and English questions from the train partition (3 answerable, 2 unanswerable cases).\footnote{We leave experiments with other proportions of answerable and unanswerable few-shot examples to future work.}
We use the same examples for all models and questions.

As \textbf{evaluation metrics}, we use precision (P), recall (R), and F1-score (F).
To evaluate sentence-level retrieval (i.e. the binary task of selecting a sentence as an answer to the question), metrics are first computed per question at the sentence level 
and then macro-averaged over questions. 

We also separately evaluate the binary task of identifying questions as \textbf{unanswerable} given the context.
Here all metrics are at the question level.
We consider two setups for extracting ``unanswerable question'' predictions from models: In the \textit{inferred} setup, we run the models as before and treat generated empty lists (in the case of LLMs) or all-zero-vectors (in the case of DeBERTa) as classifying the question as unanswerable. In the \textit{explicit} setup, we change the LLM instructions and classifier architecture to make an explicit binary prediction for each question.

During experimentation and hyperparameter selection, we evaluated only on the development split of OMoS-QA (results in \cref{app:dev-results}). Here we report our main results on the test split with the hyperparameters found during development.

\subsection{Evaluated Models}

We focus on open-weight models from MistralAI and Meta:
Mixtral-8x7B (introduced in \cref{sec:question-generation}), Mistral-7B-Instruct-v0.2 \citep[Mistral-7B;][]{jiang2023mistral} as well as Llama-3-8B-Instruct (Llama-3-8B) and Llama-3-70B-Instruct (Llama-3-70B) which are both successors of the Llama 2 model family \citep{touvron2023llama}.
We access these models via HuggingFace.\footnote{\url{https://huggingface.co}}
For comparison, we include results of the closed-source GPT-3.5-Turbo-0125 (GPT-3.5-Turbo) by OpenAI.\footnote{\url{https://platform.openai.com/docs/models/gpt-3-5-turbo}}

\subsection{Baseline}

As a baseline, we run a sentence-wise classifier, consisting of a pretrained DeBERTa-v3-large encoder \citep[][accessed via HuggingFace]{he2021deberta} and a binary classification head.\footnote{The head is a linear layer with 1024 input and 2 output features on top of a pooling layer. Additional hyperparameters are given in \cref{tab:finetuning-hyperparameters}.}
For each sentence in a document, we pass the following input to the model:
\texttt{\orange{[CLS]} <question> \purple{[SEP]} <context> \darkgreen{[SEN]} <target sentence> \darkgreen{[SEN]} <context> \purple{[SEP]}}, where the classification is made based on the encoding of the \texttt{\orange{[CLS]}} token, the target sentence is surrounded by three context sentences on the left and right (altogether surrounded by \texttt{\purple{[SEP]}} tokens),\footnote{The context size of 3 has been determined via experimentation on the dev set.} and we add the new \texttt{\darkgreen{[SEN]}} special token to the vocabulary to mark the target sentence.

We finetune the full model on OMoS-QA.

\subsection{Results}\label{sec:pretrained}

We present our results in the left half of \cref{tab:qa-experiment-test}.
All models show very good precision (70--90\%), with the highest numbers achieved by the Llama-3 models. Recall is much lower in general, with a wider span across models, reaching as low as 20.3\% (Mistral-7B 5-shot in German). Most models reach between 40\% and 50\% recall while maintaining high precision, which seems to be a favorable trade-off.
Keep in mind that selecting fewer but clearly relevant sentences, as opposed to more noisy ones, is generally in line with our goals of providing trustworthy results.
The highest precision is achieved by Mistral-7B -shot for German and GPT-3.5-Turbo 0-shot for English.

\mbox{Mixtral-8x7B}, \mbox{Llama-3-70B}, and DeBERTa strike the best overall precision/recall trade-offs (F1-score). DeBERTa in particular has almost equal precision and recall.

The last row of the table presents an approximation of the \enquote{human performance} as measured via the inter-annotator agreement (F1-score) in our dataset.
For each question, the data labeled by the various annotators is assigned to one of two sets randomly, and then one set is treated as the gold standard and the other as the system.
As the German and English versions of the dataset consist of the same (potentially translated, see \cref{sec:final-data}) questions and documents, the score is the same in the two languages.
We provide a version of this human score before majority voting and including adjacent sentences, which gives an idea of the difficulty of the task, even for humans---though note that these are untrained voluntary annotators and trained experts might achieve higher agreement.
Due to the data mismatch, this is not directly comparable to the system evaluation setup, thus we also provide a more optimistic version after filtering (``test partition only''), which is computed on the same data as the models.

\paragraph{Identifying Unanswerable Questions.}

We report results separately for the subset of questions where the human annotators agreed that the answer is not in the text (right side of \cref{tab:qa-experiment-test}).
Here, recall reflects how many of the unanswerable questions were correctly identified by the model as such.
Precision indicates how many of the questions predicted as unanswerable
did indeed not have an answer in the provided text.

For identifying unanswerable questions, we put higher priority on recall over precision, in line with our cautious approach to a sensitive scenario.
And indeed we find that overall, recall is higher and precision lower than in sentence extraction. In many cases, recall is higher than precision.

In \cref{tab:explicit-unanswerability-test} we see that explicitly instructing or training models to recognize unanswerable questions has different effects depending on the model type. Changing the zero-shot prompt given to Llama-3-70B increases recall and decreases precision compared to inferring this decision from an empty prediction. Changing the training task of the DeBERTa-classifier has the opposite effect. This might be a result of the decrease in the amount of training data that DeBERTa receives---only one example per question in the explicit setting versus one example per document sentence per question in the inferred setting. This quantitative difference does not apply to the LLM, which instead profits from the more precisely-phrased prompt.

\paragraph{Zero-shot vs. Few-shot.}

\begin{table}[t]
    \setlength{\tabcolsep}{4.5pt}
    \centering
    \footnotesize
    \begin{tabular}{ll | ccc | ccc}
\toprule
& & \multicolumn{3}{c|}{German} & \multicolumn{3}{c}{English} \\
 \multicolumn{2}{l|}{Model \& Method} & P & R & F & P & R & F \\
\midrule
 Llama-3 & Exp & 59.0 & \textbf{83.6} & \textbf{69.2} & 62.3 & \textbf{78.2} & \textbf{69.4} \\
 70B & Inf & 69.8 & 67.3 & 68.5 & 74.5 & 63.6 & 68.6 \\
\midrule
 DeBERTa & Exp & \textbf{75.0} & 43.6 & 55.2 & \textbf{75.0} & 54.5 & 63.2 \\
 & Inf & 56.2 & 65.5 & 60.5 & 59.4 & 69.1 & 63.9 \\
\bottomrule
    \end{tabular}
    \caption{Test set performance (in \%) of zero-shot Llama-3-70B and finetuned DeBERTa on explicit and inferred question-level unanswerability detection. The best result in each column is \textbf{bolded}. Exp=Explicit, Inf=Inferred.}
    \label{tab:explicit-unanswerability-test}
\end{table}

In most conditions, few-shot learning from 5 examples is beneficial either for both recall and precision, or for recall without hurting precision too much.
An exception is Mistral-7B, which overshoots on extracting fewer answers in the 5-shot scenario, with a strongly increased recall on unanswerable questions, but a worse performance on the answerable questions.

\begin{figure}[t]
    \centering
    \includegraphics[width=\columnwidth]{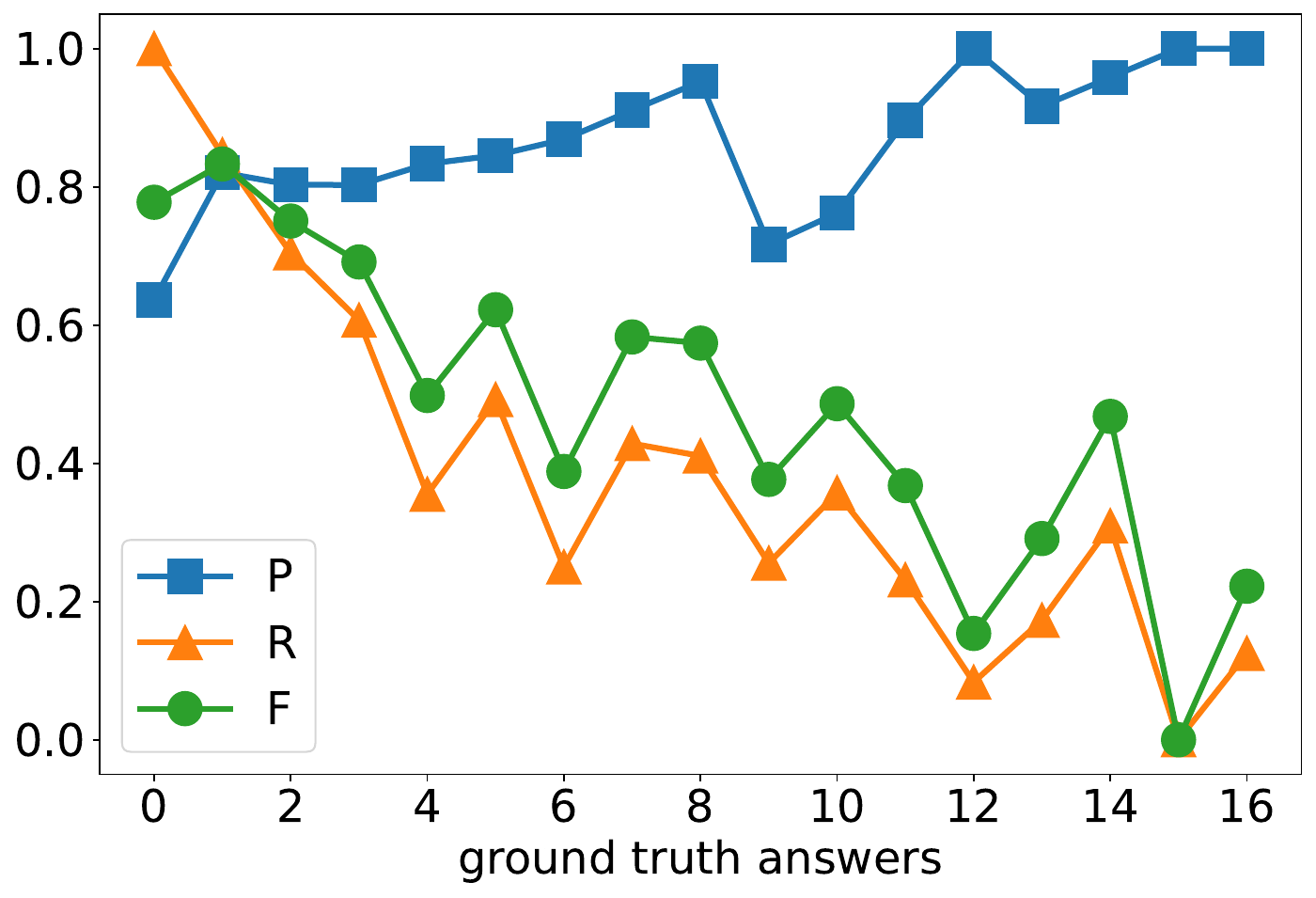}
    \caption{Test set performance as a function of the number of ground-truth answer sentences (0-shot Llama3-70B on German questions and documents).}
    \label{fig:precision-recall}
\end{figure}

\paragraph{Performance by Number of Answer Sentences.}

In all conditions and metrics (P, R, F) we observe standard deviations over individual datapoints (questions with at least one ground-truth answer) between $\pm$ 30 and $\pm$ 40 metric points. This variance can in part be explained by the varying difficulty of questions with increasing numbers of ground-truth answer sentences. The average number of gold answer sentences (henceforth~\enquote{\#answers}) lies between 5 and 6 in German and around 4 in English (\cref{tab:corpus-stats}). We show model performance as a function of \#answers exemplarily for one German model in \cref{fig:precision-recall}. As can be expected, average recall becomes roughly linearly more difficult as \#answers increases, whereas average precision already starts high and approaches 1.0 for questions with more than 10 annotated answer sentences.

\begin{table}[t]
    \setlength{\tabcolsep}{10pt}
    \centering
    \small
    \begin{tabular}{ll | cc | cc}
\toprule
       &         & \multicolumn{2}{c|}{Answerable} & \multicolumn{2}{c}{Unanswerable} \\
 Doc. & Q. & P & R & P & R \\
\midrule
 Ger. & Ger. & 85.5 & 46.6 & 69.8 & 67.3 \\
 Ger. & Eng. & \textbf{85.8} & \textbf{48.2} & 70.9 & 70.9 \\
 Ger. & Ara. & 84.6 & 41.8 & 63.1 & \textbf{74.5} \\
\midrule
 Eng. & Ara. & 80.6 & 44.0 & 74.0 & 67.3 \\
 Eng. & Eng. & 84.8 & 46.7 & \textbf{74.5} & 63.6 \\
 Eng. & Ger. & 83.2 & 45.6 & 73.5 & 65.5 \\
\midrule
 Ara. & Ara. & 80.9 & 42.2 & 71.4 & 54.5 \\
 Ara. & Eng. & 82.7 & 44.4 & 74.0 & 67.3 \\
 Ara. & Ger. & 81.9 & 43.1 & 72.7 & 72.7 \\
\bottomrule
    \end{tabular}
    \caption{Test set performance (\%) of 0-shot Llama-3-70B on \textbf{cross-language} question-context pairs.}
    \label{tab:crossling-stats}
\end{table}

\subsection{Cross-language QA}\label{sec:cross-lang}

We also conduct a pilot cross-language QA study with German, English, and Arabic questions and documents. We compare scenarios where the question language does not match the document language against scenarios where it does.
We choose Llama-3-70B over Mixtral-8x7B for this experiment, because while both perform well in \cref{sec:pretrained}, the latter was used to generate our questions.

Our findings are shown in \cref{tab:crossling-stats}. Surprisingly, asking a question in a different language than the document does not hurt performance by a lot. In fact, it seems that asking questions in English works best, regardless of document language, and German documents work best, regardless of question language.

\begin{table}[t]
    \setlength{\tabcolsep}{4.3pt}
    \centering\small
    \begin{tabular}{lcc}
    \toprule
        Model & Context toks & Params  \\
              &  (Thousands) & (Billions) \\ \midrule
       Mixtral-8x7B  & 32\footnotemark[13] &\phantom{.(12.9)} 46.7 (12.9)\footnotemark[13] \\
       Mistral-7B    & 32\footnotemark[14] &  7\footnotemark[14] \\
       Llama-3-8B    & \phantom{0}8\footnotemark[15] &  8\footnotemark[15] \\
       Llama-3-70B   &  \phantom{0}8\footnotemark[15] &  70\footnotemark[15]\phantom{0} \\
       GPT-3.5-Turbo  &  16\footnotemark[16] &  unknown \\
       DeBERTa-v3-large    &  \phantom{0}1\footnotemark[17] &  \phantom{00}0.4\footnotemark[17]\phantom{.} \\
       \bottomrule
    \end{tabular}
    \caption{Model sizes. Mixtral has a total of 46.7B parameters but uses only a subset of 12.9 of them for each token.}
    \label{tab:model_sizes}
\end{table}

\footnotetext[13]{\url{https://mistral.ai/news/mixtral-of-experts/}}
\footnotetext[14]{\url{https://huggingface.co/mistralai/Mistral-7B-Instruct-v0.2}}
\footnotetext[15]{\url{https://huggingface.co/meta-llama/Meta-Llama-3-8B-Instruct}}
\footnotetext[16]{\url{https://platform.openai.com/docs/models/gpt-3-5-turbo}}
\footnotetext[17]{\url{https://huggingface.co/microsoft/deberta-v3-large}}

\section{Discussion}\label{sec:discussion}

We interpret our results as largely positive, in particular with respect to our goal of building a reliable system that errs on the side of presenting fewer, higher precision results to the user.
On our dataset, the newest open-weight models Mixtral-8x7B and Llama-3-70B can easily compete with closed-weight GPT-3.5.

With our various evaluation criteria  and prompting setups (0-shot vs. 5-shot), we highlight different models' individual strengths: 
For example, the smaller LLMs Mistral-7B and Llama-3-8B are best at selectively identifying high-confidence answer sentences only, leading to extremely high sentence precision and unanswerability recall. They might thus lend themselves to an answerability filtering step, after which other models like Mixtral-8x7B and Llama-3-70B can do the heavy-lifting of higher-recall answer extraction.

It is important to keep in mind that we already use Mixtral-8x7B to generate questions, which likely contributes to its good performance \citep[cf.][]{panickssery2024llm}.

Our cross-language QA experiment suggests that translating questions asked in lower-resource languages (such as Arabic) to English and performing QA on German documents is a promising approach.
\cref{app:multiling} provides additional experimental results with translated and back-translated questions, which suggest that automatic translation is useful for Arabic and Ukrainian, but not so much for French, which is more similar to German and English in terms of both data availability and grammar.
In future experiments, it will be interesting to introduce additional noise into questions before prompting, such as spelling errors or code-mixing, to simulate realistic user interactions and measure models' robustness.

While LLMs are indeed powerful and flexible tools that can be quickly adapted to a specialized task via in-context learning from few-shot prompts, we also see that the best-performing LLMs in our setting are the ones with the most parameters (\cref{tab:model_sizes}). Much smaller, specialized models, such as task-specific classifiers built upon DeBERTa or other BERT-style encoders, are generally more controllable, interpretable, and environmentally friendly. Together with the competitive QA performance in terms of F1 and well-balanced precision and recall we observe, this emphasizes that this model class is still very much viable for practical applications in sensitive scenarios.

We will take these findings into account as we continuously work towards automating the document retrieval component and a service-ready implementation of the full QA system, and including more and more languages as potential query and document languages.

\section{Conclusion}
In this paper, we address the task of providing high-precision, knowledge-grounded answers to users who have freshly immigrated to Germany. 
We approach this challenge by compiling, manually annotating, and filtering a novel dataset, OMoS-QA, containing in total 900 document-question pairs in German and English. The dataset will be available to the research community under a CC-BY license.
We also present experimental results on our new dataset from a comparison of 5 LLMs and a finetuned classifier, as well as a pilot cross-language QA study.
Our results are promising and open the doors to future finetuning and large-scale multilingual experiments.

\section*{Limitations}
The OMoS-QA dataset is designed to support extractive QA in an online counseling system for immigrants.
In this paper, we have modeled an admittedly simplified scenario in which the document (potentially) containing the answer to a question is already provided (an assumption that is made in most currently used QA benchmarks).
A full search scenario would of course also require identifying potentially relevant documents, i.e., include a search component.

Another limitation of our work is that annotators were not trained specifically for our task. We counterbalance this issue by double-annotations and extensive filtering.

Finally, the current version of OMoS-QA is limited to German and English documents and questions.
As immigrants arrive from all over the world, an in particular in urgent crises without the possibility to study German in advance, more work is necessary to mitigate the language barrier. In future work, we plan to also conduct experiments for an extended set of languages.

\section*{Ethics Statement}
During dataset construction, annotators participated on a voluntary basis and agreed to the anonymized publishing of their annotations.
Before starting the annotations, they agreed to the terms shown in \cref{apx:annotation-tool}.
As the annotation study only included marking relevant answers to technical questions in text, i.e., annotators did not have to write text or provide personal information, no IRB review was deemed necessary.

Online migration counseling offers convenience and accessibility, but it also comes with several challenges.\footnote{This list was compiled with the help of ChatGPT, yet it reflects our own opinion as well.}
First of all, there is a lack of a personal connection, which may be crucial in our scenario.
Ensuring client confidentiality can be more challenging in an online environment.
Misinterpretation of cultural cues or nuances in communication may occur, leading to misunderstandings or ineffective counseling outcomes.
Finally, there are also technological barriers: not everyone has access to reliable internet connections or appropriate devices.
Yet, our work is a first attempt at developing reliable language technology to support the immigration counseling process.
Municipalities could, for example, provide computer terminals at the immigration authorities' offices, townhalls, or libraries.
And being able to search for information in a targeted system is still much of an advantage compared to waiting for an appointment for weeks.
Moreover, such a system would also lead to a more effective use of the official counselor's time, as it would relieve them from providing advice in \enquote{easy} cases.

\section*{Acknowledgments}

We are extremely grateful to our voluntary annotators for their hard work and to Tür an Tür Digitalfabrik GmbH, in particular Daniel Kehne und Sven Seeberg-Elverfeldt, for supporting this work and funding the first author. We also thank Alexander Knapp and members of the Augsburg HLT group for inspiring discussions, as well as the anonymous reviewers for their helpful comments and suggestions.

\bibliography{anthology,custom}

\begin{thebibliography}{48}
\expandafter\ifx\csname natexlab\endcsname\relax\def\natexlab#1{#1}\fi

\bibitem[{Agarwal et~al.(2022)Agarwal, Gite, Laddha, Bhattacharyya, Kar, Ekbal, Thind, Zele, and Shankar}]{agarwal-etal-2022-knowledge}
Ankush Agarwal, Raj Gite, Shreya Laddha, Pushpak Bhattacharyya, Satyanarayan Kar, Asif Ekbal, Prabhjit Thind, Rajesh Zele, and Ravi Shankar. 2022.
\newblock \href {https://aclanthology.org/2022.lrec-1.673} {Knowledge graph - deep learning: A case study in question answering in aviation safety domain}.
\newblock In \emph{Proceedings of the Thirteenth Language Resources and Evaluation Conference}, pages 6260--6270, Marseille, France. European Language Resources Association.

\bibitem[{Alfter et~al.(2023)Alfter, Volodina, Fran{\c{c}}ois, J{\"o}nsson, and Rennes}]{nlp4call-2023-natural}
David Alfter, Elena Volodina, Thomas Fran{\c{c}}ois, Arne J{\"o}nsson, and Evelina Rennes, editors. 2023.
\newblock \href {https://aclanthology.org/2023.nlp4call-1.0} {\emph{Proceedings of the 12th Workshop on NLP for Computer Assisted Language Learning}}. LiU Electronic Press, T{\'o}rshavn, Faroe Islands.

\bibitem[{Bechet et~al.(2022)Bechet, Antoine, Auguste, and Damnati}]{bechet-etal-2022-question}
Frederic Bechet, Elie Antoine, J{\'e}r{\'e}my Auguste, and G{\'e}raldine Damnati. 2022.
\newblock \href {https://aclanthology.org/2022.lrec-1.486} {Question generation and answering for exploring digital humanities collections}.
\newblock In \emph{Proceedings of the Thirteenth Language Resources and Evaluation Conference}, pages 4561--4568, Marseille, France. European Language Resources Association.

\bibitem[{Beese et~al.(2022)Beese, P{\"u}tz, and Eger}]{beese-etal-2022-fairger}
Dominik Beese, Ole P{\"u}tz, and Steffen Eger. 2022.
\newblock \href {https://arxiv.org/abs/2210.04359} {{F}air{G}er: Using {NLP} to measure support for women and migrants in 155 years of {G}erman parliamentary debates}.
\newblock ArXiv preprint arXiv:2210.04359.

\bibitem[{Blätte et~al.(2020)Blätte, Gehlhar, and Leonhardt}]{blaette-etal-2020-europeanization1}
Andreas Blätte, Simon Gehlhar, and Christoph Leonhardt. 2020.
\newblock \href {https://aclanthology.org/2020.parlaclarin-1.12} {The {E}uropeanization of parliamentary debates on migration in {A}ustria, {F}rance, {G}ermany, and the {N}etherlands}.
\newblock In \emph{Proceedings of the Second ParlaCLARIN Workshop}, pages 66--74, Marseille, France. European Language Resources Association.

\bibitem[{Brabant et~al.(2022)Brabant, Lecorv{\'e}, and Rojas~Barahona}]{brabant-etal-2022-coqar}
Quentin Brabant, Gw{\'e}nol{\'e} Lecorv{\'e}, and Lina~M. Rojas~Barahona. 2022.
\newblock \href {https://aclanthology.org/2022.lrec-1.13} {{C}o{QAR}: Question rewriting on {C}o{QA}}.
\newblock In \emph{Proceedings of the Thirteenth Language Resources and Evaluation Conference}, pages 119--126, Marseille, France. European Language Resources Association.

\bibitem[{Brook~Weiss et~al.(2021)Brook~Weiss, Roit, Klein, Ernst, and Dagan}]{brook-weiss-etal-2021-qa}
Daniela Brook~Weiss, Paul Roit, Ayal Klein, Ori Ernst, and Ido Dagan. 2021.
\newblock \href {https://doi.org/10.18653/v1/2021.emnlp-main.778} {{QA}-align: Representing cross-text content overlap by aligning question-answer propositions}.
\newblock In \emph{Proceedings of the 2021 Conference on Empirical Methods in Natural Language Processing}, pages 9879--9894, Online and Punta Cana, Dominican Republic. Association for Computational Linguistics.

\bibitem[{Charlet et~al.(2020)Charlet, Damnati, Bechet, Marzinotto, and Heinecke}]{charlet-etal-2020-cross}
Delphine Charlet, Geraldine Damnati, Frederic Bechet, Gabriel Marzinotto, and Johannes Heinecke. 2020.
\newblock \href {https://aclanthology.org/2020.lrec-1.674} {Cross-lingual and cross-domain evaluation of machine reading comprehension with squad and {CALOR}-quest corpora}.
\newblock In \emph{Proceedings of the Twelfth Language Resources and Evaluation Conference}, pages 5491--5497, Marseille, France. European Language Resources Association.

\bibitem[{Clark and Gardner(2018)}]{clark-gardner-2018-simple}
Christopher Clark and Matt Gardner. 2018.
\newblock \href {https://doi.org/10.18653/v1/P18-1078} {Simple and effective multi-paragraph reading comprehension}.
\newblock In \emph{Proceedings of the 56th Annual Meeting of the Association for Computational Linguistics (Volume 1: Long Papers)}, pages 845--855, Melbourne, Australia. Association for Computational Linguistics.

\bibitem[{Dugan et~al.(2022)Dugan, Miltsakaki, Upadhyay, Ginsberg, Gonzalez, Choi, Yuan, and Callison-Burch}]{dugan-etal-2022-feasibility}
Liam Dugan, Eleni Miltsakaki, Shriyash Upadhyay, Etan Ginsberg, Hannah Gonzalez, DaHyeon Choi, Chuning Yuan, and Chris Callison-Burch. 2022.
\newblock \href {https://doi.org/10.18653/v1/2022.findings-acl.151} {A feasibility study of answer-agnostic question generation for education}.
\newblock In \emph{Findings of the Association for Computational Linguistics: ACL 2022}, pages 1919--1926, Dublin, Ireland. Association for Computational Linguistics.

\bibitem[{Han et~al.(2022)Han, Castro~Ferreira, and Gardent}]{han-etal-2022-generating}
Kelvin Han, Thiago Castro~Ferreira, and Claire Gardent. 2022.
\newblock \href {https://aclanthology.org/2022.lrec-1.29} {Generating questions from {W}ikidata triples}.
\newblock In \emph{Proceedings of the Thirteenth Language Resources and Evaluation Conference}, pages 277--290, Marseille, France. European Language Resources Association.

\bibitem[{He et~al.(2015)He, Lewis, and Zettlemoyer}]{he-etal-2015-question}
Luheng He, Mike Lewis, and Luke Zettlemoyer. 2015.
\newblock \href {https://doi.org/10.18653/v1/D15-1076} {Question-answer driven semantic role labeling: Using natural language to annotate natural language}.
\newblock In \emph{Proceedings of the 2015 Conference on Empirical Methods in Natural Language Processing}, pages 643--653, Lisbon, Portugal. Association for Computational Linguistics.

\bibitem[{He et~al.(2021)He, Liu, Gao, and Chen}]{he2021deberta}
Pengcheng He, Xiaodong Liu, Jianfeng Gao, and Weizhu Chen. 2021.
\newblock \href {https://openreview.net/forum?id=XPZIaotutsD} {{\{}DEBERTA{\}}: {\{}DECODING{\}}-{\{}enhanced{\}} {\{}bert{\}} {\{}with{\}} {\{}disentangled{\}} {\{}attention{\}}}.
\newblock In \emph{International Conference on Learning Representations}.

\bibitem[{Henning et~al.(2023)Henning, Anthonio, Zhou, Adel, Mesgar, and Friedrich}]{henning-etal-2023-answer}
Sophie Henning, Talita Anthonio, Wei Zhou, Heike Adel, Mohsen Mesgar, and Annemarie Friedrich. 2023.
\newblock \href {https://aclanthology.org/2023.findings-emnlp.949} {Is the answer in the text? challenging {C}hat{GPT} with evidence retrieval from instructive text}.
\newblock In \emph{Findings of the Association for Computational Linguistics: EMNLP 2023}, pages 14229--14241, Singapore. Association for Computational Linguistics.

\bibitem[{Hu et~al.(2018)Hu, Peng, Huang, Qiu, Wei, and Zhou}]{hu-etal-2018-reinforced}
Minghao Hu, Yuxing Peng, Zhen Huang, Xipeng Qiu, Furu Wei, and Ming Zhou. 2018.
\newblock Reinforced mnemonic reader for machine reading comprehension.
\newblock In \emph{Proceedings of the 27th International Joint Conference on Artificial Intelligence}, IJCAI'18, page 4099–4106. AAAI Press.

\bibitem[{Ji et~al.(2023)Ji, Lee, Frieske, Yu, Su, Xu, Ishii, Bang, Madotto, and Fung}]{ji-etal-2023-hallucination}
Ziwei Ji, Nayeon Lee, Rita Frieske, Tiezheng Yu, Dan Su, Yan Xu, Etsuko Ishii, Ye~Jin Bang, Andrea Madotto, and Pascale Fung. 2023.
\newblock \href {https://doi.org/10.1145/3571730} {Survey of hallucination in natural language generation}.
\newblock \emph{ACM Comput. Surv.}, 55(12).

\bibitem[{Jiang et~al.(2023)Jiang, Sablayrolles, Mensch, Bamford, Chaplot, de~las Casas, Bressand, Lengyel, Lample, Saulnier, Lavaud, Lachaux, Stock, Scao, Lavril, Wang, Lacroix, and Sayed}]{jiang2023mistral}
Albert~Q. Jiang, Alexandre Sablayrolles, Arthur Mensch, Chris Bamford, Devendra~Singh Chaplot, Diego de~las Casas, Florian Bressand, Gianna Lengyel, Guillaume Lample, Lucile Saulnier, Lélio~Renard Lavaud, Marie-Anne Lachaux, Pierre Stock, Teven~Le Scao, Thibaut Lavril, Thomas Wang, Timothée Lacroix, and William~El Sayed. 2023.
\newblock \href {http://arxiv.org/abs/2310.06825} {Mistral 7b}.
\newblock ArXiv Preprint 2310.06825.

\bibitem[{Jiang et~al.(2024)Jiang, Sablayrolles, Roux, Mensch, Savary, Bamford, Chaplot, de~las Casas, Hanna, Bressand, Lengyel, Bour, Lample, Lavaud, Saulnier, Lachaux, Stock, Subramanian, Yang, Antoniak, Scao, Gervet, Lavril, Wang, Lacroix, and Sayed}]{jiang2024mixtral}
Albert~Q. Jiang, Alexandre Sablayrolles, Antoine Roux, Arthur Mensch, Blanche Savary, Chris Bamford, Devendra~Singh Chaplot, Diego de~las Casas, Emma~Bou Hanna, Florian Bressand, Gianna Lengyel, Guillaume Bour, Guillaume Lample, Lélio~Renard Lavaud, Lucile Saulnier, Marie-Anne Lachaux, Pierre Stock, Sandeep Subramanian, Sophia Yang, Szymon Antoniak, Teven~Le Scao, Théophile Gervet, Thibaut Lavril, Thomas Wang, Timothée Lacroix, and William~El Sayed. 2024.
\newblock \href {http://arxiv.org/abs/2401.04088} {Mixtral of experts}.

\bibitem[{Kochmar et~al.(2023)Kochmar, Burstein, Horbach, Laarmann-Quante, Madnani, Tack, Yaneva, Yuan, and Zesch}]{bea-2023-innovative}
Ekaterina Kochmar, Jill Burstein, Andrea Horbach, Ronja Laarmann-Quante, Nitin Madnani, Ana{\"\i}s Tack, Victoria Yaneva, Zheng Yuan, and Torsten Zesch, editors. 2023.
\newblock \href {https://aclanthology.org/2023.bea-1.0} {\emph{Proceedings of the 18th Workshop on Innovative Use of NLP for Building Educational Applications (BEA 2023)}}. Association for Computational Linguistics, Toronto, Canada.

\bibitem[{Lapesa et~al.(2020)Lapesa, Blessing, Blokker, Dayanik, Haunss, Kuhn, and Pad{\'o}}]{lapesa-etal-2020-debatenet}
Gabriella Lapesa, Andre Blessing, Nico Blokker, Erenay Dayanik, Sebastian Haunss, Jonas Kuhn, and Sebastian Pad{\'o}. 2020.
\newblock \href {https://aclanthology.org/2020.lrec-1.115} {{DE}bate{N}et-mig15:tracing the 2015 immigration debate in {G}ermany over time}.
\newblock In \emph{Proceedings of the Twelfth Language Resources and Evaluation Conference}, pages 919--927, Marseille, France. European Language Resources Association.

\bibitem[{Lauriola et~al.(2022)Lauriola, Small, and Moschitti}]{lauriola-etal-2022-building}
Ivano Lauriola, Kevin Small, and Alessandro Moschitti. 2022.
\newblock \href {https://aclanthology.org/2022.lrec-1.502} {Building a dataset for automatically learning to detect questions requiring clarification}.
\newblock In \emph{Proceedings of the Thirteenth Language Resources and Evaluation Conference}, pages 4701--4707, Marseille, France. European Language Resources Association.

\bibitem[{Liu et~al.(2020)Liu, Ott, Goyal, Du, Joshi, Chen, Levy, Lewis, Zettlemoyer, and Stoyanov}]{liu-etal-2020-roberta}
Yinhan Liu, Myle Ott, Naman Goyal, Jingfei Du, Mandar Joshi, Danqi Chen, Omer Levy, Mike Lewis, Luke Zettlemoyer, and Veselin Stoyanov. 2020.
\newblock \href {https://openreview.net/forum?id=SyxS0T4tvS} {Ro{\{}bert{\}}a: A robustly optimized {\{}bert{\}} pretraining approach}.

\bibitem[{Luo et~al.(2022)Luo, Hashimoto, Yavuz, Liu, Baral, and Zhou}]{luo-etal-2022-choose}
Man Luo, Kazuma Hashimoto, Semih Yavuz, Zhiwei Liu, Chitta Baral, and Yingbo Zhou. 2022.
\newblock \href {https://doi.org/10.18653/v1/2022.spanlp-1.2} {Choose your {QA} model wisely: A systematic study of generative and extractive readers for question answering}.
\newblock In \emph{Proceedings of the 1st Workshop on Semiparametric Methods in NLP: Decoupling Logic from Knowledge}, pages 7--22, Dublin, Ireland and Online. Association for Computational Linguistics.

\bibitem[{Luthier and Popescu-Belis(2020)}]{luthier-popescu-belis-2020-chat}
Gabriel Luthier and Andrei Popescu-Belis. 2020.
\newblock \href {https://aclanthology.org/2020.lrec-1.672} {Chat or learn: a data-driven robust question-answering system}.
\newblock In \emph{Proceedings of the Twelfth Language Resources and Evaluation Conference}, pages 5474--5480, Marseille, France. European Language Resources Association.

\bibitem[{McKenna et~al.(2023)McKenna, Li, Cheng, Hosseini, Johnson, and Steedman}]{mckenna-etal-2023-sources}
Nick McKenna, Tianyi Li, Liang Cheng, Mohammad Hosseini, Mark Johnson, and Mark Steedman. 2023.
\newblock \href {https://doi.org/10.18653/v1/2023.findings-emnlp.182} {Sources of hallucination by large language models on inference tasks}.
\newblock In \emph{Findings of the Association for Computational Linguistics: EMNLP 2023}, pages 2758--2774, Singapore. Association for Computational Linguistics.

\bibitem[{M{\"o}ller et~al.(2021)M{\"o}ller, Risch, and Pietsch}]{moller-etal-2021-germanquad}
Timo M{\"o}ller, Julian Risch, and Malte Pietsch. 2021.
\newblock \href {https://doi.org/10.18653/v1/2021.mrqa-1.4} {{G}erman{Q}u{AD} and {G}erman{DPR}: Improving non-{E}nglish question answering and passage retrieval}.
\newblock In \emph{Proceedings of the 3rd Workshop on Machine Reading for Question Answering}, pages 42--50, Punta Cana, Dominican Republic. Association for Computational Linguistics.

\bibitem[{Murdock et~al.(2012)Murdock, Fan, Lally, Shima, and Boguraev}]{murdock-etal-2012-textual}
J.~W. Murdock, J.~Fan, A.~Lally, H.~Shima, and B.~K. Boguraev. 2012.
\newblock \href {https://doi.org/10.1147/JRD.2012.2187249} {Textual evidence gathering and analysis}.
\newblock \emph{IBM Journal of Research and Development}, 56(3.4):8:1--8:14.

\bibitem[{Narayan et~al.(2018)Narayan, Cardenas, Papasarantopoulos, Cohen, Lapata, Yu, and Chang}]{narayan-etal-2018-document}
Shashi Narayan, Ronald Cardenas, Nikos Papasarantopoulos, Shay~B. Cohen, Mirella Lapata, Jiangsheng Yu, and Yi~Chang. 2018.
\newblock \href {https://doi.org/10.18653/v1/P18-1188} {Document modeling with external attention for sentence extraction}.
\newblock In \emph{Proceedings of the 56th Annual Meeting of the Association for Computational Linguistics (Volume 1: Long Papers)}, pages 2020--2030, Melbourne, Australia. Association for Computational Linguistics.

\bibitem[{Panickssery et~al.(2024)Panickssery, Bowman, and Feng}]{panickssery2024llm}
Arjun Panickssery, Samuel~R. Bowman, and Shi Feng. 2024.
\newblock \href {http://arxiv.org/abs/2404.13076} {Llm evaluators recognize and favor their own generations}.
\newblock ArXiv Preprint 2404.13076.

\bibitem[{Pouran Ben~Veyseh et~al.(2022)Pouran Ben~Veyseh, Lai, Dernoncourt, and Nguyen}]{pouran-ben-veyseh-etal-2022-behanceqa}
Amir Pouran Ben~Veyseh, Viet Lai, Franck Dernoncourt, and Thien Nguyen. 2022.
\newblock \href {https://aclanthology.org/2022.lrec-1.796} {{B}ehance{QA}: A new dataset for identifying question-answer pairs in video transcripts}.
\newblock In \emph{Proceedings of the Thirteenth Language Resources and Evaluation Conference}, pages 7321--7327, Marseille, France. European Language Resources Association.

\bibitem[{Prasad et~al.(2023)Prasad, Bui, Yoon, Deilamsalehy, Dernoncourt, and Bansal}]{prasad-etal-2023-meetingqa}
Archiki Prasad, Trung Bui, Seunghyun Yoon, Hanieh Deilamsalehy, Franck Dernoncourt, and Mohit Bansal. 2023.
\newblock \href {https://doi.org/10.18653/v1/2023.acl-long.837} {{M}eeting{QA}: Extractive question-answering on meeting transcripts}.
\newblock In \emph{Proceedings of the 61st Annual Meeting of the Association for Computational Linguistics (Volume 1: Long Papers)}, pages 15000--15025, Toronto, Canada. Association for Computational Linguistics.

\bibitem[{Radford et~al.(2018)Radford, Narasimhan, Salimans, Sutskever et~al.}]{radford-etal-2018-improving}
Alec Radford, Karthik Narasimhan, Tim Salimans, Ilya Sutskever, et~al. 2018.
\newblock \href {https://s3-us-west-2.amazonaws.com/openai-assets/research-covers/language-unsupervised/language_understanding_paper.pdf} {Improving language understanding by generative pre-training}.

\bibitem[{Rajpurkar et~al.(2018)Rajpurkar, Jia, and Liang}]{rajpurkar-etal-2018-know}
Pranav Rajpurkar, Robin Jia, and Percy Liang. 2018.
\newblock \href {https://doi.org/10.18653/v1/P18-2124} {Know what you don{'}t know: Unanswerable questions for {SQ}u{AD}}.
\newblock In \emph{Proceedings of the 56th Annual Meeting of the Association for Computational Linguistics (Volume 2: Short Papers)}, pages 784--789, Melbourne, Australia. Association for Computational Linguistics.

\bibitem[{Rajpurkar et~al.(2016)Rajpurkar, Zhang, Lopyrev, and Liang}]{rajpurkar-etal-2016-squad}
Pranav Rajpurkar, Jian Zhang, Konstantin Lopyrev, and Percy Liang. 2016.
\newblock \href {https://doi.org/10.18653/v1/D16-1264} {{SQ}u{AD}: 100,000+ questions for machine comprehension of text}.
\newblock In \emph{Proceedings of the 2016 Conference on Empirical Methods in Natural Language Processing}, pages 2383--2392, Austin, Texas. Association for Computational Linguistics.

\bibitem[{Roit et~al.(2020)Roit, Klein, Stepanov, Mamou, Michael, Stanovsky, Zettlemoyer, and Dagan}]{roit-etal-2020-controlled}
Paul Roit, Ayal Klein, Daniela Stepanov, Jonathan Mamou, Julian Michael, Gabriel Stanovsky, Luke Zettlemoyer, and Ido Dagan. 2020.
\newblock \href {https://doi.org/10.18653/v1/2020.acl-main.626} {Controlled crowdsourcing for high-quality {QA}-{SRL} annotation}.
\newblock In \emph{Proceedings of the 58th Annual Meeting of the Association for Computational Linguistics}, pages 7008--7013, Online. Association for Computational Linguistics.

\bibitem[{Ross et~al.(2016)Ross, Rist, Carbonell, Cabrera, Kurowsky, and Wojatzki}]{ross-etal-2016-measuring}
Björn Ross, Michael Rist, Guillermo Carbonell, Benjamin Cabrera, Nils Kurowsky, and Michael Wojatzki. 2016.
\newblock \href {https://arxiv.org/abs/1701.08118} {Measuring the reliability of hate speech annotations: The case of the {E}uropean refugee crisis}.
\newblock In \emph{Proceedings of the 3rd Workshop on Natural Language Processing for Computer-mediated Communication (NLP4CMC)}, number~17 in Bochumer linguistische Arbeitsberichte: BLA, pages 6--9.

\bibitem[{Sanguinetti et~al.(2018)Sanguinetti, Poletto, Bosco, Patti, and Stranisci}]{sanguinetti-etal-2018-italian}
Manuela Sanguinetti, Fabio Poletto, Cristina Bosco, Viviana Patti, and Marco Stranisci. 2018.
\newblock \href {https://aclanthology.org/L18-1443} {An {I}talian {T}witter corpus of hate speech against immigrants}.
\newblock In \emph{Proceedings of the Eleventh International Conference on Language Resources and Evaluation ({LREC} 2018)}, Miyazaki, Japan. European Language Resources Association (ELRA).

\bibitem[{Schick and Sch{\"u}tze(2021)}]{schick-schutze-2021-generating}
Timo Schick and Hinrich Sch{\"u}tze. 2021.
\newblock \href {https://doi.org/10.18653/v1/2021.emnlp-main.555} {Generating datasets with pretrained language models}.
\newblock In \emph{Proceedings of the 2021 Conference on Empirical Methods in Natural Language Processing}, pages 6943--6951, Online and Punta Cana, Dominican Republic. Association for Computational Linguistics.

\bibitem[{Seo et~al.(2017)Seo, Kembhavi, Farhadi, and Hajishirzi}]{seo-etal-2017-bidirectional}
Minjoon Seo, Aniruddha Kembhavi, Ali Farhadi, and Hannaneh Hajishirzi. 2017.
\newblock \href {https://openreview.net/forum?id=HJ0UKP9ge} {Bidirectional attention flow for machine comprehension}.
\newblock In \emph{International Conference on Learning Representations}.

\bibitem[{Shah and Bender(2024)}]{shah-bender-2024-envisioning}
Chirag Shah and Emily~M. Bender. 2024.
\newblock \href {https://doi.org/10.1145/3649468} {Envisioning information access systems: What makes for good tools and a healthy web?}
\newblock \emph{ACM Trans. Web}.

\bibitem[{Touvron et~al.(2023)Touvron, Martin, Stone, Albert, Almahairi, Babaei, Bashlykov, Batra, Bhargava, Bhosale, Bikel, Blecher, Ferrer, Chen, Cucurull, Esiobu, Fernandes, Fu, Fu, Fuller, Gao, Goswami, Goyal, Hartshorn, Hosseini, Hou, Inan, Kardas, Kerkez, Khabsa, Kloumann, Korenev, Koura, Lachaux, Lavril, Lee, Liskovich, Lu, Mao, Martinet, Mihaylov, Mishra, Molybog, Nie, Poulton, Reizenstein, Rungta, Saladi, Schelten, Silva, Smith, Subramanian, Tan, Tang, Taylor, Williams, Kuan, Xu, Yan, Zarov, Zhang, Fan, Kambadur, Narang, Rodriguez, Stojnic, Edunov, and Scialom}]{touvron2023llama}
Hugo Touvron, Louis Martin, Kevin Stone, Peter Albert, Amjad Almahairi, Yasmine Babaei, Nikolay Bashlykov, Soumya Batra, Prajjwal Bhargava, Shruti Bhosale, Dan Bikel, Lukas Blecher, Cristian~Canton Ferrer, Moya Chen, Guillem Cucurull, David Esiobu, Jude Fernandes, Jeremy Fu, Wenyin Fu, Brian Fuller, Cynthia Gao, Vedanuj Goswami, Naman Goyal, Anthony Hartshorn, Saghar Hosseini, Rui Hou, Hakan Inan, Marcin Kardas, Viktor Kerkez, Madian Khabsa, Isabel Kloumann, Artem Korenev, Punit~Singh Koura, Marie-Anne Lachaux, Thibaut Lavril, Jenya Lee, Diana Liskovich, Yinghai Lu, Yuning Mao, Xavier Martinet, Todor Mihaylov, Pushkar Mishra, Igor Molybog, Yixin Nie, Andrew Poulton, Jeremy Reizenstein, Rashi Rungta, Kalyan Saladi, Alan Schelten, Ruan Silva, Eric~Michael Smith, Ranjan Subramanian, Xiaoqing~Ellen Tan, Binh Tang, Ross Taylor, Adina Williams, Jian~Xiang Kuan, Puxin Xu, Zheng Yan, Iliyan Zarov, Yuchen Zhang, Angela Fan, Melanie Kambadur, Sharan Narang, Aurelien Rodriguez, Robert Stojnic, Sergey Edunov, and Thomas
  Scialom. 2023.
\newblock \href {http://arxiv.org/abs/2307.09288} {Llama 2: Open foundation and fine-tuned chat models}.
\newblock ArXiv Preprint 2307.09288.

\bibitem[{Wang et~al.(2019)Wang, Yu, Sun, Chen, Yu, McAllester, and Roth}]{wang-etal-2019-evidence}
Hai Wang, Dian Yu, Kai Sun, Jianshu Chen, Dong Yu, David McAllester, and Dan Roth. 2019.
\newblock \href {https://doi.org/10.18653/v1/K19-1065} {Evidence sentence extraction for machine reading comprehension}.
\newblock In \emph{Proceedings of the 23rd Conference on Computational Natural Language Learning (CoNLL)}, pages 696--707, Hong Kong, China. Association for Computational Linguistics.

\bibitem[{Wang(2024)}]{wang-2024-metaphorical}
Yunxiao Wang. 2024.
\newblock \href {https://aclanthology.org/2024.latechclfl-1.3} {Metaphorical framing of refugees, asylum seekers and immigrants in {UK}s left and right-wing media}.
\newblock In \emph{Proceedings of the 8th Joint SIGHUM Workshop on Computational Linguistics for Cultural Heritage, Social Sciences, Humanities and Literature (LaTeCH-CLfL 2024)}, pages 18--27, St. Julians, Malta. Association for Computational Linguistics.

\bibitem[{Westera et~al.(2020)Westera, Mayol, and Rohde}]{westera-etal-2020-ted}
Matthijs Westera, Laia Mayol, and Hannah Rohde. 2020.
\newblock \href {https://aclanthology.org/2020.lrec-1.141} {{TED}-{Q}: {TED} talks and the questions they evoke}.
\newblock In \emph{Proceedings of the Twelfth Language Resources and Evaluation Conference}, pages 1118--1127, Marseille, France. European Language Resources Association.

\bibitem[{Xu et~al.(2022)Xu, Ouyang, and Liu}]{xu-etal-2022-task}
Zhuoqun Xu, Liubo Ouyang, and Yang Liu. 2022.
\newblock \href {https://aclanthology.org/2022.lrec-1.670} {Task-driven and experience-based question answering corpus for in-home robot application in the {H}ouse3{D} virtual environment}.
\newblock In \emph{Proceedings of the Thirteenth Language Resources and Evaluation Conference}, pages 6232--6239, Marseille, France. European Language Resources Association.

\bibitem[{Yoon et~al.(2020)Yoon, Dernoncourt, Kim, Bui, and Jung}]{yoon-etal-2020-propagate}
Seunghyun Yoon, Franck Dernoncourt, Doo~Soon Kim, Trung Bui, and Kyomin Jung. 2020.
\newblock \href {https://aclanthology.org/2020.lrec-1.664} {Propagate-selector: Detecting supporting sentences for question answering via graph neural networks}.
\newblock In \emph{Proceedings of the Twelfth Language Resources and Evaluation Conference}, pages 5400--5407, Marseille, France. European Language Resources Association.

\bibitem[{Yuan et~al.(2023)Yuan, Wang, Wang, Fine, Abdelghani, Sauz{\'e}on, and Oudeyer}]{yuan-etal-2023-selecting}
Xingdi Yuan, Tong Wang, Yen-Hsiang Wang, Emery Fine, Rania Abdelghani, H{\'e}l{\`e}ne Sauz{\'e}on, and Pierre-Yves Oudeyer. 2023.
\newblock \href {https://doi.org/10.18653/v1/2023.findings-acl.820} {Selecting better samples from pre-trained {LLM}s: A case study on question generation}.
\newblock In \emph{Findings of the Association for Computational Linguistics: ACL 2023}, pages 12952--12965, Toronto, Canada. Association for Computational Linguistics.

\bibitem[{Zwitter~Vitez et~al.(2022)Zwitter~Vitez, Brglez, Robnik~{\v{S}}ikonja, {\v{S}}kvorc, Vezovnik, and Pollak}]{zwitter-vitez-etal-2022-extracting}
Ana Zwitter~Vitez, Mojca Brglez, Marko Robnik~{\v{S}}ikonja, Tadej {\v{S}}kvorc, Andreja Vezovnik, and Senja Pollak. 2022.
\newblock \href {https://aclanthology.org/2022.lrec-1.259} {Extracting and analysing metaphors in migration media discourse: towards a metaphor annotation scheme}.
\newblock In \emph{Proceedings of the Thirteenth Language Resources and Evaluation Conference}, pages 2430--2439, Marseille, France. European Language Resources Association.

\end{thebibliography}

\clearpage
\appendix

\section{Chance-corrected Jaccard Coefficient}\label{app:agreement-metric}

For computing agreement, we use a chance-corrected version of the Jaccard coefficient.
For a question $q_i$, it is defined as follows for two sets of selected answer sentences $A_{i_a} \subseteq S_i$ and $A_{i_b} \subseteq S_i$, where $S_i$ is the set of all sentences of the document, and $a$ and $b$ index the two annotators:

\[agr_{obs} = J(A_{i_a}, A_{i_b}) 
= \frac{|A_{i_a} \cap A_{i_b}|}{|A_{i_a} \cup A_{i_b}|} 
\]

For $A_{i_a} = A_{i_b} = \emptyset$ we set $J(A_{i_a}, A_{i_b}) = 1$ as both annotators completely agree that there is no answer.

\textbf{Chance Correction.}
In order to account for the possibility of authors just agreeing \enquote{by chance,} chance correction can be applied. As the prior probability $P(sel)$ of a sentence $s_{i_k} \in S_i$ being selected we take the amount of all sentence selection in the whole corpus divided by the amount of all sentences in the corpus times 2 to account for two annotations being made:

\[P(sel) = \frac{\sum_{i = 1}^n (|A_{i_a}| + |A_{i_b}|)}{2 * \sum_{i = 1}^n |S_i|}\]

The probability $P(agr)$ that two random annotations agree on a sentence being an answer is then:

\[agr_{exp} = P(agr) = P(sel)^2\]

$P(agr)$ is therefore the expected agreement $agr_{exp}$.
The observed agreement $agr_{obs}$ is the Jaccard index $J(A_{i_a}, A_{i_b})$, such that the chance-corrected Jaccard index can be calculated as follows:
\[J_{cc}(A_{i_a}, A_{i_b}) 
= \frac{agr_{obs} - agr_{exp}}{1 - agr_{exp}} 
\]

\section{Prompt Template}\label{apx:prompt-template}

As mentioned in \cref{sec:experiment-setup} we mostly follow the prompt template proposed by \citet{henning-etal-2023-answer} for both our 0-shot and 5-shot experiments. As Mixtral-8x7B and Mistral-7B do not support messages with the \emph{system} role, we only include \emph{user} and \emph{assistant} messages for these models. Our complete 0-shot prompt:
\\ \textbf{system}: \texttt{Your task is to select sentences from a document that answer a given question.} (Llama-3 models and GPT-3.5-Turbo only)
\\ \textbf{user (question, document)}: \texttt{Given the question and document below, select the sentences from the document that answer the question.
It may also be the case that none of the sentences answers the question.
In the document, each sentence is marked with an ID.
Output the IDs of the relevant sentences as a list, e.g., ``[1,2,3]'', and output ``[]'' if no sentence is relevant.
Output only these lists.
\\ Question: \{question\}
\\ Document: \{document\}}

We use the chunked samples shown in \cref{fig:prompt-samples} (or their sentence-by-sentence translations) for the 5-shot experiments. 
For each sample we insert the following two messages to the prompt before the final user message:
\\ \textbf{user (question, document)}
\\ \textbf{assistant (answers)}: \texttt{\{answers\}}

\begin{figure*}
    \small
    \textbf{Question 1}: \texttt{What do you need to open a bank account?}
    \\ \textbf{Document 1}:
    \\ \texttt{[9] When can I start learning to drive?}
    \\ \texttt{[10] In Germany, you may only drive a car with a valid driver\'s license.}
    \\ \texttt{[11] Beforehand, you have to attend a driving school and take theoretical and practical lessons, which you also have to pay for.}
    \\ \texttt{[12] You can get information about this at the driving school.}
    \\ \texttt{[13] When can I open my own bank account?}
    \\ \textbf{Answer 1}: \texttt{[]}
    \\
    \\ \textbf{Question 2}: \texttt{What is a fictitious certificate?}
    \\ \textbf{Document 2}:
    \\ \texttt{[0] Residence with fictitious certificate}
    \\ \texttt{[1] Departure with a fictitious certificate}
    \\ \texttt{[2] With a fictitious certificate, you have a temporary right of residence.}
    \\ \texttt{[3] There are different types of fictitious certificate.}
    \\ \texttt{[4]Please note:}
    \\ \texttt{[5] Re-entry into the federal territory is only possible with a fictitious certificate in accordance with § 81 para.4 AufenthG possible.}
    \\ \textbf{Answer 2}: \texttt{[2]}
    \\
    \\ \textbf{Question 3}: \texttt{Where can I find information on admission procedures at vocational schools?}
    \\ \textbf{Document 3}:
    \\ \texttt{[11] Initial vocational training is possible at vocational schools and vocational colleges.}
    \\ \texttt{[12] Training can take place both in the dual system (training company and vocational school) or ``purely'' school-based training (vocational schools).}
    \\ \texttt{[13] The dates and registration requirements vary from vocational school to vocational school.}
    \\ \texttt{[14] Information evenings are held at vocational schools every year before enrollment.}
    \\ \texttt{[15] Information on the admission procedure at the vocational schools can be obtained directly from the respective school.}
    \\ \texttt{[5] Re-entry into the federal territory is only possible with a fictitious certificate in accordance with § 81 para.4 AufenthG possible.}
    \\ \textbf{Answer 3}: \texttt{[14, 15]}
    \\
    \\ \textbf{Question 4}: \texttt{What types of school are there in Germany?}
    \\ \textbf{Document 4}:
    \\ \texttt{[0] Support with school or personal problems}
    \\ \texttt{[1] Does your child need help with problems?}
    \\ \texttt{[2] Then these places will help you:}
    \\ \texttt{[3] Youth social work (JaS for short) and youth work at schools (JA for short) for school, personal or family problems:}
    \\ \texttt{[4] It is best to contact the school directly or the Augsburg District Office for general information:}
    \\ \textbf{Answer 4}: \texttt{[0]}
    \\
    \\ \textbf{Question 5}: \texttt{What topics are covered in the initial orientation courses?}
    \\ \textbf{Document 5}:
    \\ \texttt{[2] The German courses for initial language orientation (also known as initial orientation courses) teach both basic German language skills and information about life in Germany.}
    \\ \texttt{[3] They are a practical starting aid in the new living environment and make everyday life easier.}
    \\ \texttt{[4] A course comprises 300 teaching units of 45 minutes each and covers topics such as ``Health/medical care'', ``Work'', ``Kindergarten/school'', ``Housing'', ``Local orientation/transport/mobility''.}
    \\ \texttt{[5] The focus is on oral communication: participants should learn as quickly as possible to find their way around in everyday life.}
    \\ \texttt{[6] Across all modules, initial orientation courses are also about teaching values.}
    \\ \textbf{Answer 5}: \texttt{[2, 4, 5, 6]}
    \caption{Chunked samples for 5-shot experiments.}
    \label{fig:prompt-samples}
\end{figure*}

\section{Custom Annotation Tool}\label{apx:annotation-tool}

\begin{figure*}
    \centering
    \includegraphics[width=0.9\textwidth]{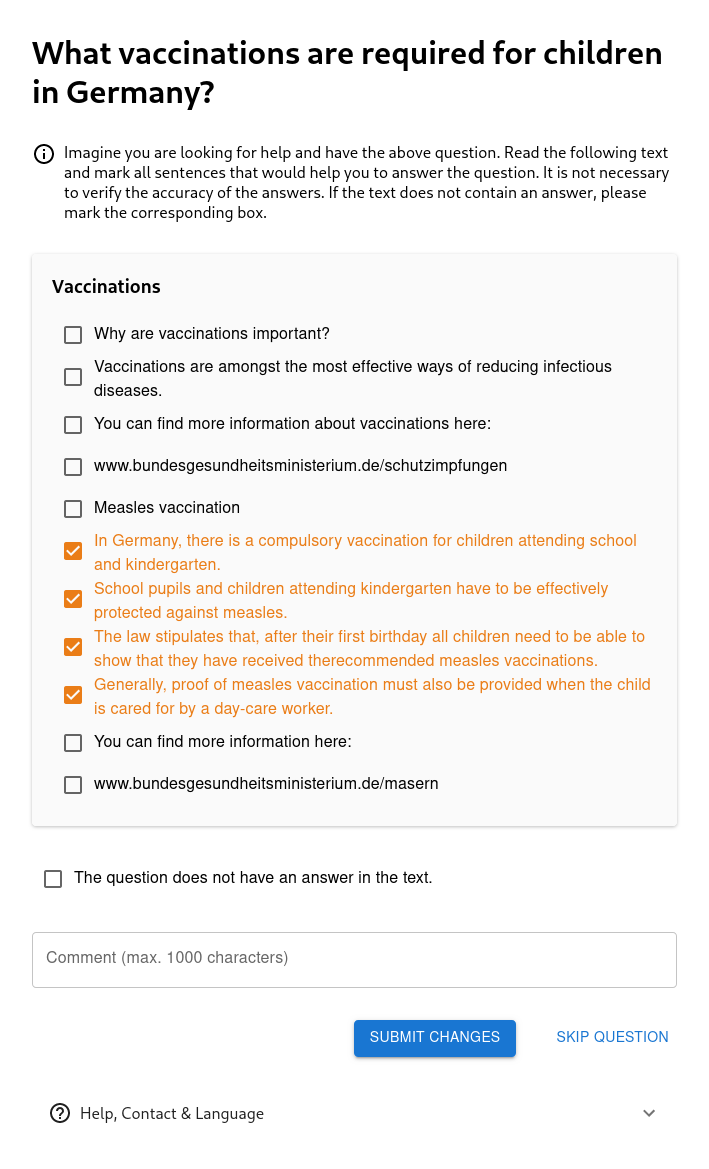}
    \caption{Custom annotation tool.}
    \label{fig:annotation-tool}
\end{figure*}

For the human annotations described in \cref{sec:human-annotations} we developed a custom web-based annotation tool for the selection of the answer sentences. All human annotators agreed to the following conditions: \emph{I agree to the processing and publication of my annotations and their use for machine learning. All annotations and information entered will be stored and processed anonymously.}
\cref{fig:annotation-tool} shows a screenshot of the custom annotation tool.

\section{Development Set Performance}
\label{app:dev-results}

\begin{table*}[t]
    \centering
    \footnotesize
    \begin{tabular}{l | c | c}
\toprule
& Sentence Classification & Question Classification \\
\midrule
Batch size & 8 & 8 \\
Learning rate & $2 * 10^{-6}$ & $2 * 10^{-6}$ \\
Weight decay & 0.1 & 0.1 \\
Warmup steps & 50 & 50 \\
Evaluation steps & 50 & 10 \\
Max. epochs & 3 & 10 \\
Early stopping & 10 & 10 \\
\bottomrule
    \end{tabular}
    \caption[Hyperparameters for finetuning DeBERTa.]{The used hyperparameters for finetuning DeBERTa for answer extraction using binary sentence classification and question answerability classification.}
    \label{tab:finetuning-hyperparameters}
\end{table*}

We observe slightly different trends on the development set (\cref{tab:qa-experiment-dev}) than on the test set (\cref{tab:qa-experiment-test}).
Namely, three 0-shot model setups have a particularly low recall on sentence extraction:
Mixtral-8x7B, Llama-3-8B, and GPT-3.5, which means in conjunction with high precision that they tend to generally extract fewer sentences per question. Out of these three, Mixtral-8x7B and Llama-3-8B also have particularly low precision at identifying unanswerable questions, meaning that more often than not they do not extract any answer sentence for questions which would in fact be answerable given the context. This gets largely fixed by providing few-shot examples.

\begin{table*}[t]
    \centering
    \footnotesize
    \begin{tabular}{ll | ccc | ccc c ccc | ccc}
\toprule
\multicolumn{2}{c}{} & \multicolumn{6}{c}{Answerable questions: sentence-level} & & \multicolumn{6}{c}{Identifying unanswerable questions} \\\cmidrule{1-8}\cmidrule{10-15} 
       &         & \multicolumn{3}{c|}{German} & \multicolumn{3}{c}{English}  & &  \multicolumn{3}{c|}{German} & \multicolumn{3}{c}{English} \\
 Model & Setting & P & R & F & P & R & F & & P & R & F & P & R & F \\
\cmidrule{1 - 8}\cmidrule{10 - 15}
 Mixtral-8x7B & 0-shot & 74.1 & 31.7 & 32.4 & 67.7 & 29.0 & 29.3 & \textbf{} & 38.8 & 81.6 & 52.5 & 41.1 & 78.9 & 54.1 \\
  & 5-shot & 74.0 & \textbf{58.0} & \textbf{55.6} & 72.9 & \textbf{53.9} & 52.4 &  & 76.2 & 84.2 & 80.0 & 68.9 & 81.6 & 74.7 \\
\cmidrule{1 - 8}\cmidrule{10 - 15}
 Mistral-7B & 0-shot & 74.0 & 45.5 & 47.0 & 76.7 & 45.7 & 48.4 &  & 59.5 & 57.9 & 58.7 & 58.3 & 55.3 & 56.8 \\
  & 5-shot & 71.0 & 42.8 & 40.7 & 70.5 & 39.0 & 38.7 &  & 50.9 & 71.1 & 59.3 & 52.8 & 73.7 & 61.5 \\
\cmidrule{1 - 8}\cmidrule{10 - 15}
 Llama-3-8B & 0-shot & \textbf{89.8} & 26.8 & 30.0 & \textbf{86.2} & 33.9 & 37.7 &  & 32.0 & 86.8 & 46.8 & 40.8 & 81.6 & 54.4 \\
  & 5-shot & 77.6 & 44.6 & 45.8 & 78.1 & 39.2 & 40.6 &  & 61.0 & \textbf{94.7} & 74.2 & 52.3 & \textbf{89.5} & 66.0 \\
\cmidrule{1 - 8}\cmidrule{10 - 15}
 Llama-3-70B & 0-shot & 84.2 & 48.6 & 53.9 & 79.6 & 48.3 & 52.6 &  & \textbf{81.6} & 81.6 & \textbf{81.6} & \textbf{77.5} & 81.6 & \textbf{79.5} \\
  & 5-shot & 85.9 & 51.1 & 55.4 & 82.8 & 51.9 & \textbf{55.0} &  & 66.7 & 84.2 & 74.4 & 70.8 & 89.5 & 79.1 \\
\cmidrule{1 - 8}\cmidrule{10 - 15}
 GPT-3.5-Turbo & 0-shot & 70.4 & 33.9 & 38.5 & 73.3 & 36.9 & 42.6 &  & 63.2 & 63.2 & 63.2 & 73.5 & 65.8 & 69.4 \\
  & 5-shot & 77.5 & 47.9 & 50.8 & 80.7 & 44.1 & 49.5 &  & 78.9 & 78.9 & 78.9 & 71.4 & 78.9 & 75.0 \\
\cmidrule[\heavyrulewidth]{1 - 8}\cmidrule[\heavyrulewidth]{10 - 15}
\multicolumn{2}{l|}{\itshape Human Upper Bound*} & $-$ & $-$ & \textit{62.9} & $-$ & $-$ & \textit{62.9} &  & $-$ & $-$ & $-$ & $-$ & $-$ & $-$ \\
\multicolumn{2}{l|}{\textit{\hspace{2mm} with adjacent sentences}} & $-$ & $-$ & \textit{88.8} & $-$ & $-$ & \textit{88.8} &  & $-$ & $-$ & $-$ & $-$ & $-$ & $-$ \\ 
\bottomrule
    \end{tabular}
    \caption{Development set performance (in \%) of 0-shot and 5-shot LLMs on answerable questions (left) and unanswerable questions (right). The best result in each column is \textbf{bolded}. \textit{*Human upper bound} is computed from agreement data and not directly comparable.
    }
    \label{tab:qa-experiment-dev}
\end{table*}

\newpage

\section{Multilingual Experiments}\label{app:multiling}

We evaluate models on the following additional languages that are highly relevant in the migration context: Arabic (ar), French (fr), and Ukrainian (uk).
These and other languages are more challenging due to their limited resources and much different language structure (German and English are closely related).
Furthermore, Arabic and Ukrainian both use a non-Latin alphabet: The Arabic and Cyrillic alphabet.
We use machine translation with DeepL to translate the question and, sentence-by-sentence, the document for each instance of the original OMoS-QA dataset.

In order to assess possible adverse effects of leveraging machine translation and to compare it to directly querying the model with the question in its original language, we evaluate the performance in an additional retranslation setting.
To this end, we combine the original German documents with retranslated questions, i.e., questions that are first translated to the aforementioned languages and then back to German.
This corresponds to the use of machine translation in the full OMoS system, as only user input (and possibly the answers) are subject to translation, while the document corpus remains unchanged.
However, questions are translated twice in the retranslation setting and results should thus be considered as lower performance boundary.
Since German is the original dataset language of OMoS-QA, there are no results for the retranslated setting.

\begin{table*}[t]
    \setlength{\tabcolsep}{5.5pt}
    \centering
    \footnotesize
    \begin{tabular}{lc | ccc | ccc c ccc | ccc}
\toprule
\multicolumn{2}{c}{} & \multicolumn{6}{c}{Sentence-level Answers} & & \multicolumn{6}{c}{Question-level Unanswerability} \\
\cmidrule{1-8}\cmidrule{10-15} & & \multicolumn{3}{c|}{Multilingual} & \multicolumn{3}{c}{German Retrans.} & & \multicolumn{3}{c|}{Multilingual} & \multicolumn{3}{c}{German Retrans.} \\
 Model & Lang. & P & R & F & P & R & F &  & P & R & F & P & R & F \\
\cmidrule{1 - 8}\cmidrule{10 - 15}
 Mixtral-8x7B & de & 74.5 & 47.1 & 57.7 & $-$ & $-$ & $-$ &  & 68.9 & 56.4 & 62.0 & $-$ & $-$ & $-$ \\
  & ar & 72.5 & 42.7 & 53.8 & 77.8 & 45.2 & 57.2 &  & 62.8 & 49.1 & 55.1 & 55.4 & 56.4 & 55.9 \\
  & fr & 74.2 & 43.7 & 55.0 & 75.0 & 45.2 & 56.4 &  & 64.1 & 45.5 & 53.2 & 57.4 & 49.1 & 52.9 \\
  & uk & 69.3 & 46.4 & 55.6 & 74.7 & 45.8 & 56.8 &  & 73.2 & 54.5 & 62.5 & 58.2 & 58.2 & 58.2 \\
\cmidrule{1 - 8}\cmidrule{10 - 15}
 Llama-3-70B & de & \textbf{85.5} & 46.6 & 60.3 & $-$ & $-$ & $-$ &  & 69.8 & 67.3 & 68.5 & $-$ & $-$ & $-$ \\
  & ar & 80.9 & 42.2 & 55.5 & \textbf{86.0} & 44.1 & 58.3 &  & 71.4 & 54.5 & 61.9 & 61.0 & 65.5 & 63.2 \\
  & fr & 84.1 & 44.9 & 58.5 & 84.3 & 43.5 & 57.4 &  & 72.9 & 63.6 & 68.0 & 63.8 & \textbf{67.3} & 65.5 \\
  & uk & 82.4 & 41.3 & 55.0 & 85.6 & 43.3 & 57.5 &  & \textbf{74.5} & 63.6 & \textbf{68.6} & \textbf{64.9} & \textbf{67.3} & \textbf{66.1} \\
\cmidrule{1 - 8}\cmidrule{10 - 15}
 DeBERTa & de & 62.6 & \textbf{62.4} & \textbf{62.5} & $-$ & $-$ & $-$ &  & 56.2 & 65.5 & 60.5 & $-$ & $-$ & $-$ \\
  & ar & 63.3 & 54.9 & 58.8 & 65.2 & 53.5 & 58.8 &  & 43.4 & 60.0 & 50.4 & 44.0 & \textbf{67.3} & 53.2 \\
  & fr & 66.3 & 56.9 & 61.2 & 61.4 & \textbf{59.9} & \textbf{60.6} &  & 50.7 & 67.3 & 57.8 & 53.8 & 63.6 & 58.3 \\
  & uk & 54.7 & 61.4 & 57.9 & 62.2 & 55.9 & 58.8 &  & 57.1 & \textbf{72.7} & 64.0 & 48.7 & \textbf{67.3} & 56.5 \\
\bottomrule
    \end{tabular}
    \caption{Test set performance (in \%) of zero-shot LLMs and finetuned DeBERTa on sentence-level answer extraction (left) and detection of unanswerable questions (right) for multilingual and retranslated settings. In the multilingual setting, questions and documents are machine translated to the respective language. In the retranslated setting, the question is retranslated back to German and paired with the original German document. The best result in each column is \textbf{bolded}.}
    \label{tab:multilingual-experiment}
\end{table*}

\subsection{Sentence-Level Results}\label{sec:multilingual-answers}

The results are shown in \cref{tab:multilingual-experiment}.
On the left side of the table, we compare sentence-level results of different languages in both a multilingual and a retranslated setting for select models.
Compared to the performances on the original German dataset version, all models display lower performance in both the multilingual and the retranslated setting for Arabic, French, and Ukrainian.
Llama-3-70B shows slightly higher precision for retranslated Arabic ($+$0.5\%) and Ukrainian ($+$0.1\%), however, this comes at a cost of a clearer decrease in recall ($-$2.5\% and $-$3.3\% respectively).
For the multilingual setting, French results were the closest to German.
With exception to Mixtral-8x7B, the F1-score for French is at least 2\% higher.
Similarly, while retranslating improves F1-score performance compared to directly querying the LLM for Arabic and Ukrainian in all settings by up to $+$3.4\%, retranslating French comes at a performance loss for Llama-3-70B and DeBERTa.
Mixtral-8x7B, on the other hand, shows a performance improvement ($+$1.4\%) for retranslating French to German, although it is explicitly advertised as \enquote{fluent in French.}\footnote{\url{https://mistral.ai/technology\#models}}
The biggest performance loss is displayed by Llama-3-70B in the multilingual setting in Ukrainian ($-$5.3\%) and Arabic ($-$4.8\%).

In general, the observed performance differences are observable but not as notable as expected.
This is especially the case for Arabic and Ukrainian, as the differences in the alphabet, grammar, and language origins are significant.
While machine translation seems to have a slightly better performance for these languages, a performance deterioration compared to the original German dataset is still measurable.
However, the questions are translated twice in our setup, and, as a consequence, the actual implications should be smaller.

\subsection{Question-Level Unanswerability}\label{sec:multilingual-unanswerability}

As in \cref{sec:experiment-setup}, we infer question-level unanswerability from sentence-level answer extraction results.
If no sentence of a document is marked as answer, we treat the question as unanswerable given the document.
In contrast to question-level answer extraction, the German results are not necessarily better than those of other languages in the multilingual setting, but they always outperform the retranslated results.
Surprisingly, all models perform slightly better in the Ukrainian multilingual setting than on the original German dataset (up to $+$3.5\%, DeBERTa) and mostly considerably better than on Arabic and French (up to $+$13.6\%).
Especially Ukrainian precision is high among all models, which is in line with low precision on the sentence-level, i.e., more sentences are marked as answer.
Retranslating only yields small performance improvements for French for DeBERTa and for Arabic for all models.
Otherwise, directly querying models leads to better question-level results (up to $+$7.5\%).

\end{document}